\documentclass{article} 
\usepackage[preprint]{colm2025_conference}

\usepackage{microtype}
\usepackage{hyperref}
\usepackage{url}
\usepackage{booktabs}

\usepackage{lineno}

\usepackage{graphicx}
\usepackage{tikz}
\usepackage[edges]{forest}
\definecolor{hidden-draw}{RGB}{205, 44, 36}
\definecolor{hidden-blue}{RGB}{194,232,247}
\definecolor{hidden-orange}{RGB}{243,202,120}
\definecolor{hidden-yellow}{RGB}{242,244,193}
\definecolor{tree-level-1}{RGB}{245,20,85}
\definecolor{tree-level-2}{RGB}{246,86,118}
\definecolor{tree-level-3}{RGB}{248,177,193}
\definecolor{tree-leaf}{RGB}{176,230,198}

\definecolor{Self}{RGB}{255,0,128}
\definecolor{Ensemble}{RGB}{0,127,255}
\definecolor{Iterative}{RGB}{153,51,255}

\definecolor{exemplar1}{RGB}{136,98,148}
\definecolor{exemplar2}{RGB}{148,210,242}
\definecolor{knowledge1}{RGB}{249,219,152}
\definecolor{knowledge2}{RGB}{255,245,220}

\usepackage{subcaption}
\usepackage{pgfplots}
\pgfplotsset{compat=1.17}
\usetikzlibrary{pgfplots.groupplots}

\usepackage{microtype}
\usepackage{hyperref}
\usepackage{url}

\usepackage{enumitem}
\newenvironment{itemize*}%
 {\leftmargini=20pt\begin{itemize}%
  \setlength{\itemsep}{3pt}%
  \setlength{\parskip}{0pt}%
  }%
 {\end{itemize}}
\newenvironment{enumerate*}%
 {\begin{enumerate}%
  \setlength{\itemsep}{0pt}%
  \setlength{\parskip}{0pt}}%
 {\end{enumerate}}

\usepackage{cleveref}
\usepackage{listings}
\usepackage{titlesec}
\usepackage{multirow}
\usepackage{makecell}
\usepackage{xcolor}
\usepackage{bbding}

\usepackage{xspace}

\usepackage{graphicx}

\usepackage{pifont}
\newcommand{\cmark}{\textcolor[rgb]{0.0, 0.6, 0.0}{\ding{51}}} 
\newcommand{\xmark}{\textcolor[rgb]{0.7, 0.0, 0.0}{\ding{55}}} 
\definecolor{forestgreen}{RGB}{34,139,34}
\definecolor{my_yellow}{RGB}{255,165,0}
\newcommand{\gmark}{\textcolor{my_yellow}{\ding{51}}} 

\newcommand*{\img}[1]{%
    \raisebox{-.25\baselineskip}{%
        \includegraphics[
        height=1.25\baselineskip,
        width=1.25\baselineskip,
        keepaspectratio,
        ]{#1}%
    }%
}

\definecolor{darkblue}{rgb}{0, 0, 0.5}
\hypersetup{colorlinks=true, citecolor=darkblue, linkcolor=darkblue, urlcolor=darkblue}

\NewDocumentCommand{\heng}
{ mO{} }{\textcolor{red}{\textsuperscript{\textit{Heng}}\textsf{\textbf{\small[#1]}}}}

\NewDocumentCommand{\cheng}
{ mO{} }{\textcolor{orange}{\textsuperscript{\textit{Cheng}}\textsf{\textbf{\small[#1]}}}}

\NewDocumentCommand{\emre}
{ mO{} }{\textcolor{blue}{\textsuperscript{\textit{Emre}}\textsf{\textbf{\small[#1]}}}}

\NewDocumentCommand{\xiusi}
{ mO{} }{\textcolor{cyan}{\textsuperscript{\textit{Xiusi}}\textsf{\textbf{\small[#1]}}}}

\NewDocumentCommand{\vardhan}
{ mO{} }{\textcolor{forestgreen}{\textsuperscript{\textit{vardhan}}\textsf{\textbf{\small[#1]}}}}

\title{\img{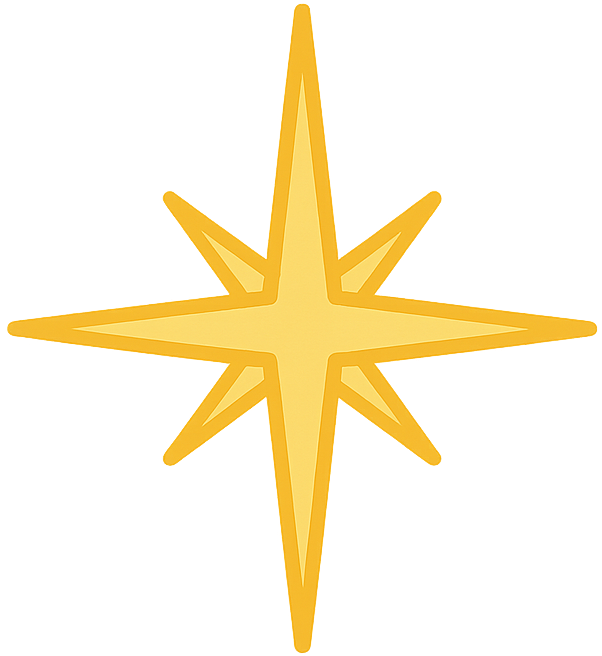}A Desideratum for Conversational Agents: Capabilities, Challenges, and Future Directions}

\author{Emre Can Acikgoz\thanks{Equal Contribution.}, \quad Cheng Qian\footnotemark[1], \quad Hongru Wang\footnotemark[1], \; \textbf{Vardhan Dongre}, \; \textbf{Xiusi Chen}, \;\\
\textbf{Heng Ji},  \quad \textbf{Dilek Hakkani-Tür}, \quad \textbf{Gokhan Tur} \\
University of Illinois Urbana-Champaign\\ 
\texttt{\{acikgoz2, chengq9, hrwise98, hengji, dilek, gokhan\}@illinois.edu} \\
}

\NewDocumentCommand{\hr}
{ mO{} }{\textcolor{purple}{\textsuperscript{\textit{Hongru}}\textsf{\textbf{\small[#1]}}}}

\begin{document}

\ifcolmsubmission
\linenumbers
\fi

\maketitle

\begin{abstract}
Recent advances in Large Language Models (LLMs) have propelled conversational AI from traditional dialogue systems into sophisticated agents capable of autonomous actions, contextual awareness, and multi-turn interactions with users. 
Yet, fundamental questions about their capabilities, limitations, and paths forward remain open.
This survey paper presents a desideratum for next-generation Conversational Agents—\textit{what has been achieved}, \textit{what challenges persist}, and \textit{what must be done for more scalable systems that approach human-level intelligence}. 
To that end, we systematically analyze LLM-driven Conversational Agents by organizing their capabilities into three primary dimensions: (i) \textbf{Reasoning}—logical, systematic thinking inspired by human intelligence for decision making, (ii) \textbf{Monitor}—encompassing self-awareness and user interaction monitoring, and (iii) \textbf{Control}—focusing on tool utilization and policy following.
Building upon this, we introduce a novel taxonomy by classifying recent work on Conversational Agents around our proposed desideratum. 
We identify critical research gaps and outline key directions, including \textbf{realistic evaluations}, \textbf{long-term multi-turn reasoning skills}, \textbf{self-evolution capabilities}, \textbf{collaborative and multi-agent task completion}, \textbf{personalization}, and \textbf{proactivity}. 
This work aims to provide a structured foundation, highlight existing limitations, and offer insights into potential future research directions for Conversational Agents, ultimately advancing progress toward Artificial General Intelligence (AGI). We maintain a curated repository of papers at: \url{https://github.com/emrecanacikgoz/awesome-conversational-agents}.

\end{abstract}

\section{Introduction}
\vspace{-3mm}

Conversational AI systems have long pursued the goal of human-like interactions~\citep{young02_icslp}.
Similarly, the ambition to develop robust AI agents with a high degree of autonomy and adaptive intelligence has also remained a central focus in the field~\citep{Minsky1986}.
Within the rapid emergence of LLMs~\citep{Achiam2023GPT4TR, Dubey2024TheL3-llama3, guo2025deepseek}, these advances have led to dialogue systems that excel at multi-turn conversations~\citep{chung-etal-2023-instructtods, hudecek-dusek-2023-are_llms_tod, feng-etal-2023-ldst, wang2023ds_survey}, while also enabling language agents to effectively leverage external tools for complex real-world tasks~\citep{parisi2022talm, schick2023toolformer, wang-etal-2023-large, qin2024toollearningfm}.

Despite these advancements, it is important to note that while ``common'' agents may also perform complex reasoning~\citep{kumar2025reasoningsurvey}, tool usage~\citep{qin2024toollearningfm, qu2025tool} and general language capabilities~\citep{sumers2024cognitive, wang2024survey, liu2025acognitive}, they typically do not engage in interactive dialogue or adapt to user intent and environmental context in real time. Moreover, effective Conversational Agents must not only manage multifaceted reasoning and tool invocation but also maintain multi-turn coherent dialogues with complex tool usage~\citep{acikgoz2025singlemodelmastermultiturn}, clarifying ambiguous user intent~\citep{andukuri2024star, dongre2024respact}, adapting to user states~\citep{jacqmin-etal-2022-follow}, and responding empathetically~\citep{rashkin-etal-2019-towards}. 

In order to fill this gap, we propose a \textit{desideratum}—a guiding vision and set of requirements for next-generation Conversational Agents around three primary dimensions: (i) \textbf{Reasoning}, encompassing logical, systematic thinking for planning and decision making; (ii) \textbf{Monitor}, covering both self-awareness and user interaction monitoring; and (iii) \textbf{Control}, focusing on tool selection, execution, and policy following. 
In light of the extensive body of work on reasoning, monitoring, and control, our work first defines a desideratum that organizes the capabilities of Conversational Agents into these three primary dimensions. While existing research has laid the groundwork by studying reasoning, monitoring, and control capabilities under our desideratum, significant challenges remain such as long-term multi-turn reasoning and policy following, self-evolution, personalization, and proactivity. We highlight these challenges and then propose a roadmap for future research for developing more capable, robust, and intelligent Conversational Agents.

\noindent\textbf{Scope and Organization.} This work presents the first comprehensive survey of Conversational Agents, examining their evolution, capabilities, and challenges, while pointing a future research roadmap.
We define Conversational Agents and outline the key desiderata (Section \ref{sec:background}). 
We then review related work, emphasizing novel technical capabilities, their alignment with the proposed desiderata, and potential challenges (Section \ref{sec:taxonomy}). 
Finally, we discuss the broader implications, setting the stage for future exploration (Section \ref{sec:future_work}).\footnote{For interested readers, we also provide an additional discussion on evaluation methods and benchmarks for Conversational Agents in \Cref{app:eval-ca}.}

\begin{figure}[t!]
    \centering
    \includegraphics[width=\textwidth]{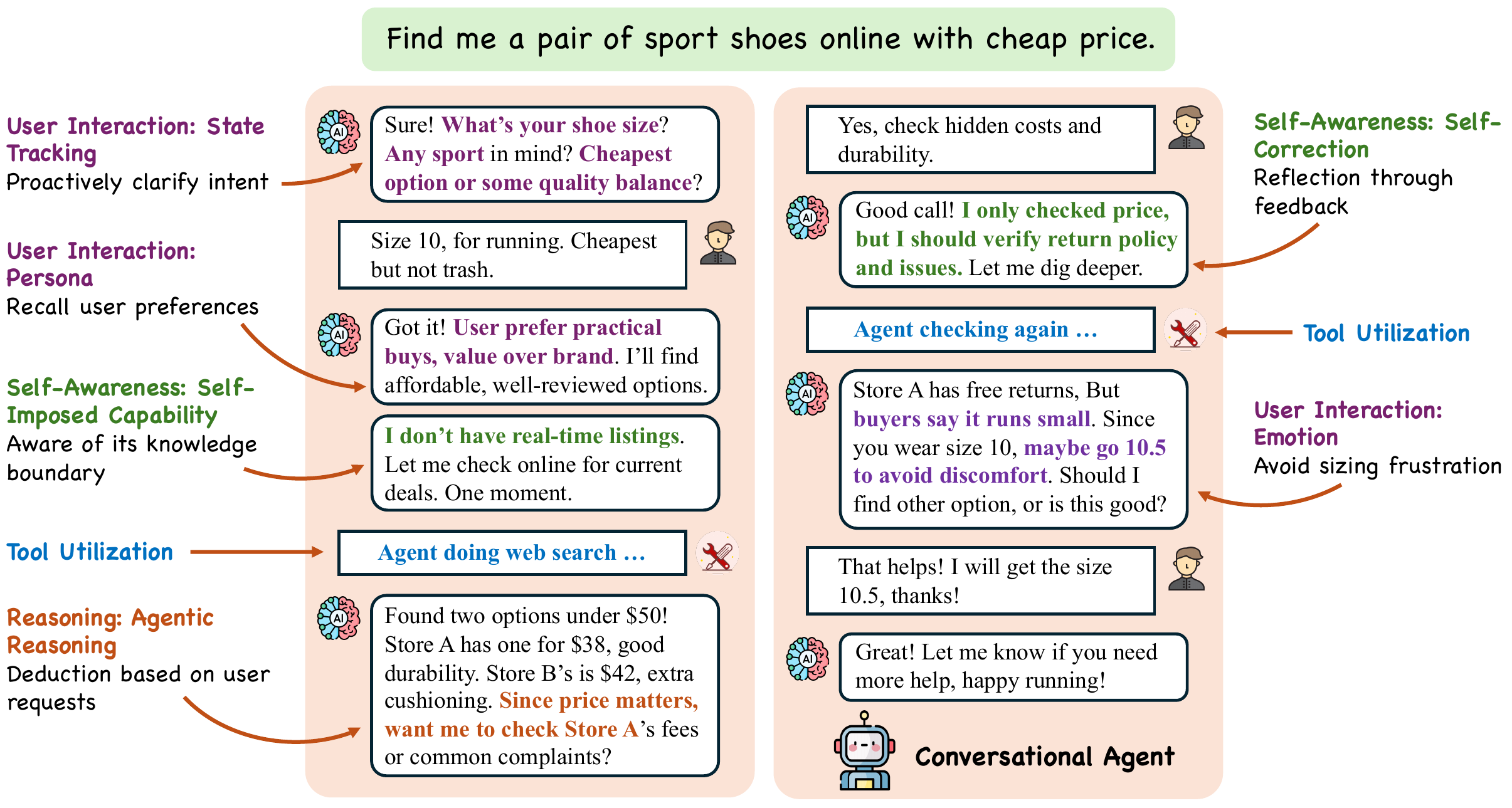}
    \caption{Overview of a Conversational Agent illustrating proposed desideratum.}
    \label{fig:overview}
    \vspace{-5mm}
\end{figure}

\vspace{-3mm}
\section{Background}
\label{sec:background}

\vspace{-3mm}

\subsection{What are Conversational Agents?}

\vspace{-2mm}

\noindent\textbf{Definition.} Traditional dialogue systems primarily focus on natural language understanding and generation for human-machine interactions \citep{chends_survey, ni2022recentadvancesdeeplearning,wang2023ds_survey}, whereas autonomous language agents emphasize decision-making and rely on tool invocation to access external knowledge sources for complex task-solving \citep{xagent2023, qin2024toolllm, wang2024survey}. Combining these strengths, a \emph{Conversational Agent} is an LLM-based framework that integrates \textbf{reasoning} to enable systematic planning and complex decision-making, leverages \textbf{monitoring} to maintain self-awareness and continuously track user interaction, and employs capabilities for \textbf{control} to adeptly utilize tools and adhere to policies (See \Cref{app:rmc}, \Cref{tab:comparison} for further discussions). 
By continuously integrating these processes across multi-turn interactions, the agent provides coherent, contextually-aware, and personalized dialogue experiences. 


\noindent\textbf{Example.} Unlike chatbots that focus primarily on response generation, Conversational Agents dynamically interpret user needs, track context during interactions, and adapt their actions while maintaining natural conversation. For instance, in \Cref{fig:overview}, when assisting a user in purchasing affordable running shoes, the agent first deduces requirements while prioritizing value over brand (reasoning). It then executes web searches to identify cost-effective options and validates store policies or hidden fees to ensure alignment with user priorities (control). Concurrently, the system tracks contextual cues such as the user’s preference for durability and adapts its approach upon discovering sizing discrepancies or feedback about return processes (monitor). By iteratively refining recommendations, the agent maintains conversational coherence, balances efficiency with user satisfaction, and self-corrects to address evolving needs, exemplifying seamless integration of reasoning, monitor, and control in multi-turn interactions.

\vspace{-3mm}
\subsection{Why do we need Conversational Agents?}
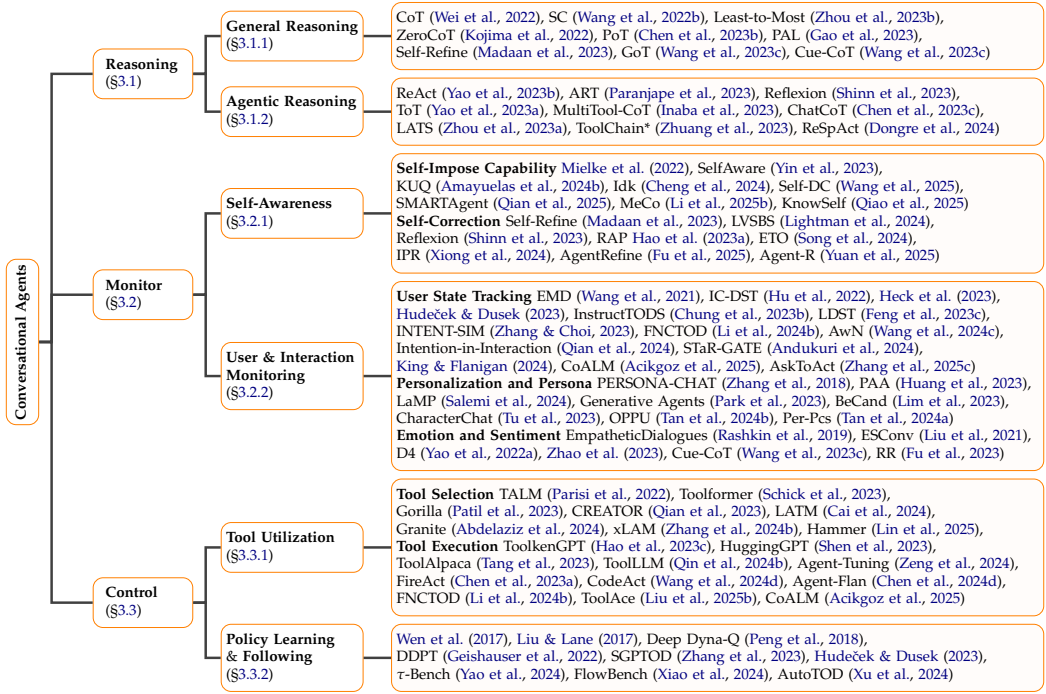
\begin{figure}[h!]
  \centering
  \resizebox{\linewidth}{!}{
        \begin{forest}
            forked edges,
            for tree={
                grow=east,
                reversed=true,
                anchor=base west,
                parent anchor=east,
                child anchor=west,
                base=left,
                font=\tiny,
                rectangle,
                draw=orange,
                rounded corners,
                align=left,
                minimum width=4em,
                edge+={darkgray, line width=1pt},
                s sep=3pt,
                inner xsep=2pt,
                inner ysep=3pt,
                ver/.style={rotate=90, child anchor=north, parent anchor=south, anchor=center},
                last level/.style={fill=orange!2}, 
            },
            where level=1{text width=3.0em,font=\tiny,}{},
            where level=2{text width=5.3em,font=\tiny,}{},
            where level=3{last level, text width=25.8em,font=\tiny,}{},
            [
                \textbf{Conversational Agents}, ver
                [
                    \textbf{Reasoning }\\ (\S \ref{sec:reason})
                    [
                        \textbf{General Reasoning} \\ (\S \ref{sec:general-reasoning})
                        [
                        CoT~\citep{wei2022chain}{,} SC~\citep{wang2022sc}{,} Least-to-Most~\citep{zhou2023leasttomost}{,} \\ ZeroCoT~\citep{kojima2022zerocot}{,} PoT~\citep{chen2023pot}{,} PAL~\citep{gao2023pal}{,} \\ Self-Refine~\citep{madaan2023sr}{,} GoT~\citep{wang-etal-2023-cue}{,} Cue-CoT~\citep{wang-etal-2023-cue}\\
                        ]
                    ]
                    [
                        \textbf{Agentic Reasoning} \\ (\S \ref{sec:agentic-reasoning})
                        [
                        ReAct~\citep{yao2023reactsynergizingreasoningacting-react}{,} ART~\citep{paranjape2023art}{,} Reflexion~\citep{shinn2023reflexion}{,}\\ ToT~\citep{yao2023tree}{,}  MultiTool-CoT~\citep{inaba2023multitool}{,} ChatCoT~\citep{chen2023chatcot}{,} \\ LATS~\citep{zhou2023lats}{,} ToolChain*~\citep{zhuang2023toolchain}{,} ReSpAct~\citep{dongre2024respact} \\ 
                        ]
                    ]
                ]
                [
                    \textbf{Monitor} \\ (\S \ref{sec:monitor})
                    [
                        \textbf{Self-Awareness} \\ (\S \ref{sec:self-awareness})
                        [
                        {
                        \textbf{Self-Impose Capability}  \cite{mielke2022reducingconvagentcalibration}{,} SelfAware~\citep{yin2023selfaware}{,}  \\ KUQ~\citep{amayuelas2024kuq}{,} Idk~\citep{cheng2024idk}{,} Self-DC~\citep{wang2025selfdcreasonactself}{,} \\ SMARTAgent~\citep{qian2025smart}{,} MeCo~\citep{li2025adaptive}, KnowSelf~\citep{qiao2025agentic} \\
                        \textbf{Self-Correction}
                        Self-Refine~\citep{madaan2023sr}{,} LVSBS~\citep{lightman2024lvsbs}{,} \\ Reflexion~\citep{shinn2023reflexion}{,} RAP~\cite{hao2023rap}{,} ETO~\citep{song2024eto}{,} \\ IPR~\citep{xiong2024ipr}{,} AgentRefine~\citep{fu2025agentrefine}{,} Agent-R~\citep{yuan2025agentr}
                        }
                        ]
                    ]
                    [
                        \textbf{User \& Interaction} \\ \textbf{Monitoring} \\ (\S \ref{sec:user-monitoring})
                        [
                        \textbf{User State Tracking} EMD~\citep{wang-etal-2021-fast}{,} IC-DST~\citep{hu2022icdst}{,} \cite{heck2023chatgptzeroshotdst}{,} \\  \cite{hudecek-dusek-2023-are_llms_tod}{,} InstructTODS~\citep{chung2023instructtods}{,} LDST~\citep{feng2023ldst}{,} \\ INTENT-SIM~\citep{zhang2023clarify}{,}  FNCTOD~\citep{li2024fnctod}{,} AwN~\citep{wang2024learning}{,} \\ Intention-in-Interaction~\citep{qian2024tell}{,} STaR-GATE~\citep{andukuri2024star}{,} \\ \cite{king-flanigan-2024-unsupervised}{,}  CoALM~\citep{acikgoz2025singlemodelmastermultiturn}{,} AskToAct~\citep{zhang2025asktoact} \\
                        \textbf{Personalization and Persona}
                         PERSONA-CHAT~\citep{zhang2018personalizingdialogue}{,} PAA~\citep{huang2023personalized}{,} \\ LaMP~\citep{salemi2024lamp}{,} Generative Agents~\citep{generative_agents}{,} BeCand~\citep{lim2023beyond}{,} \\ CharacterChat~\citep{tu2023characterchat}{,} OPPU~\citep{tan2024oppu}{,} Per-Pcs~\citep{tan2024perpcs} \\
                        \textbf{Emotion and Sentiment}
                        EmpatheticDialogues~\citep{rashkin-etal-2019-towards}{,} ESConv~\citep{liu2021towards}{,} \\ D4~\citep{yao-etal-2022-d4}{,} \cite{zhao2023chatgptemotion}{,} Cue-CoT~\citep{wang-etal-2023-cue}{,} RR~\citep{fu-etal-2023-reasoning}
                        ]
                    ]
                ]
                [
                    \textbf{Control} \\ (\S \ref{sec:control})
                    [
                        \textbf{Tool Utilization} \\ (\S \ref{sec:tool_utilization})
                        [
                        \textbf{Tool Selection} TALM~\citep{parisi2022talm}{,} Toolformer~\citep{schick2023toolformer}{,} \\ Gorilla~\citep{patil2023gorilla}{,} CREATOR~\citep{qian2023creator}{,} LATM~\citep{cai2024latm}{,} \\ Granite~\citep{abdelaziz2024granite}{,} xLAM~\citep{zhang2024xlam}{,} Hammer~\citep{lin2025hammer}{,} \\
                        \textbf{Tool Execution} ToolkenGPT~\citep{hao2023toolkengpt}{,} HuggingGPT~\citep{shen2023hugginggpt}{,} \\ ToolAlpaca~\citep{tang2023toolalpaca}{,} ToolLLM~\citep{qin2024toolllm}{,} Agent-Tuning~\citep{zeng2024agenttuning}{,} \\ FireAct~\citep{chen2023fireact}{,} CodeAct~\citep{wang2024codeact}{,} Agent-Flan~\citep{chen2024agentflan}{,} \\ FNCTOD~\citep{li2024fnctod}{,} ToolAce~\citep{liu2025toolace}{,} CoALM~\citep{acikgoz2025singlemodelmastermultiturn}\\ 
                        ]
                    ]
                    [
                        \textbf{Policy Learning} \\ \& \textbf{Following} \\ (\S \ref{sec:policy_following})
                        [
                         \cite{wen-etal-2017-network}{,} \cite{bing2017policy}{,} Deep Dyna-Q~\citep{peng-etal-2018-deep}{,} \\  DDPT~\citep{geishauser-etal-2022-dynamic}{,} SGPTOD~\citep{zhang2023sgptod}{,} \cite{hudecek-dusek-2023-are_llms_tod}{,} \\ $\tau$-Bench~\citep{Yao2024tau}{,} FlowBench~\citep{xiao2024flowbench}{,} AutoTOD~\citep{xu2024autotod} \\
                        ]
                    ]
                ]
            ]
        \end{forest}
    }
  \vspace{-5mm}
  \caption{A taxonomy of our desiderata for Conversational Agents, with representative approaches listed for each component.}
  \vspace{-3mm}
  \label{fig:taxonomy}
\end{figure}

\vspace{-2mm}
Conversational Agents, as a unified framework, combine the advantages of language agents and dialogue systems, while eliminating corresponding limitations, achieving the agentic workflow in the multi-turn conversational flow. Specifically, as user queries become more intricate, Language Agents that focus primarily on one-turn tool execution often lack the ability to track and utilize context over multiple turns with user, or traditional chatbots often struggle to invoke external tools for complex problem-solving (See \Cref{app:ca} for further details). Consequently, they both struggle with tasks such as comparing different services, booking reservations, or conducting multi-step troubleshooting, all of which require a sequence of actions rather than isolated responses. In contrast, we propose that Conversational Agents provide a universal and robust solution with the following three key features: Reasoning, Monitor, and Control, as illustrated in \Cref{fig:taxonomy}.
\vspace{-3mm}

\section{Desideratum Taxonomy for Conversational Agents}
\label{sec:taxonomy}
\vspace{-3mm}

\subsection{Reasoning}
\vspace{-2mm}
\label{sec:reason}
Reasoning equips Conversational Agents with structured decision-making capabilities, enabling them to generate coherent and contextually appropriate responses. In this section, we categorize reasoning into two dimensions: \textit{General Reasoning} and \textit{Agentic Reasoning}.


\subsubsection{General Reasoning} 
\label{sec:general-reasoning}
General reasoning methods aim to equip Conversational Agents with structured thinking capabilities that go beyond surface-level responses. Chain-of-Thought (CoT)~\citep{wei2022chain, kojima2022zerocot} introduces a step-by-step reasoning paradigm that transparently shows intermediate logic, making agents’ problem-solving more interpretable. Least-to-Most~\citep{zhou2023leasttomost} proposes decomposing complex problems into smaller sub-problems and solving them sequentially, while PAL~\citep{madaan2023sr} and PoT~\citep{chen2023pot} employ reasoning chains by combining text and code. Furthermore, Self-Consistency~\citep{wang2022sc} further refines this process by sampling multiple reasoning paths and choosing the most consistent outcome. Self-Refine~\citep{madaan2023sr} adds an iterative reflection stage, allowing the agent to revise its initial line of thought and improving the quality of final answers, whereas Cue-CoT~\citep{wang-etal-2023-cue} incorporates specialized prompts (or “cues”) to direct the reasoning chain in more targeted ways.

\subsubsection{Agentic Reasoning} 
\label{sec:agentic-reasoning}

Different from general reasoning, agentic reasoning explicitly combines structured thought processes with concrete actions, enabling Conversational Agents to not only reason internally but also interact with external environments or tools. ReAct~\citep{yao2023reactsynergizingreasoningacting-react} synergizes reasoning steps with actionable outputs by explicitly prompting the agent to deliberate before performing external actions, thus improving interpretability and practical task performance. Extending this idea, Tree of Thoughts (ToT)~\citep{yao2023tree} organizes reasoning steps into a tree structure, systematically exploring multiple reasoning paths and evaluating them to select the optimal route. Reflexion~\citep{shinn2023reflexion} enhances agentic reasoning further by enabling agents to retrospectively reflect on previous unsuccessful attempts and adjust their future strategies accordingly. On the other hand, some approaches extend CoT prompting with external tools for complex reasoning~\citep{paranjape2023art, inaba2023multitool, chen2023chatcot}, while others utilize search-based planning algorithms like MCTS to navigate expansive action spaces effectively~\citep{zhou2023lats, zhuang2023toolchain}. Unlike these approaches, ReSpAct~\citep{dongre2024respact} explicitly incorporates dynamic user interaction, enabling agents to clarify ambiguities and iteratively refine actions through user feedback, thus achieving more aligned and clear behaviors.
It is important to note that, agentic reasoning should not be confused with agentic planning (e.g., TravelPlanner~\citep{xie2024travelplanner}); reasoning involves immediate deliberation or justification of individual actions, whereas planning explicitly organizes multiple actions into coherent long-term sequences with the aim of accomplishing specific structured goals~\citep{hao2023reasoning}.

\noindent\textbf{Challenges.} These methods often rely on extensive manual prompt engineering, which is time-consuming, costly, and sometimes require specialized domain expertise. To address this challenge, some research efforts focus on automatically optimizing prompts to reduce dependency on hand-crafted solutions~\citep{khattab2023dspy, yuksekgonul2024textgrad}. However, the application of such auto-prompting approaches to agentic tasks (e.g., tool-learning) remains under-explored. Additionally, recent progress in Large Reasoning Models (LRMs) reveal the effectiveness of reinforcement learning (RL) to boost the reasoning capabilities~\citep{jaech2024o1, muennighoff2025s1, guo2025deepseek, li2025system1}. However, developing an agentic reward model (i.e., a universal verifier that provides reliable, domain-agnostic feedback) remains a major challenge~\citep{peng2025agentic}.

\subsection{Monitor}
\label{sec:monitor}

Monitoring empowers Conversational Agents with continuous awareness of their internal states and user interactions, ensuring responsive and adaptive conversations. We categorize monitoring into two dimensions: \textit{Self-Awareness} and \textit{User \& Interaction Monitoring}.
\subsubsection{Self-Awareness}
\label{sec:self-awareness}
Self-awareness enables Conversational Agents to recognize and reason about their internal states, capabilities, and limitations, which allows agents to dynamically adapt their behaviors and interactions. In this section, we discuss: \textit{Self-Knowledge Boundary}, that defines the limits of an agent’s internal knowledge; and \textit{Self-Correction}, which allows agents to refine their behavior by analyzing past mistakes, incorporating feedback, and improving over time through iterative learning.

\noindent\textbf{Self-Knowledge Boundary.} It is essential for Conversational Agents to understand the limits of their knowledge and capability~\citep{mielke2022reducingconvagentcalibration, amayuelas-etal-2024-knowledge, li2024knowledgeboundarylargelanguage}. Recognizing what lies within its scope of knowledge~\citep{yin2023selfaware, amayuelas2024kuq, cheng2024idk} and what requires external interactions enables them to make decisions more effectively and efficiently~\citep{wang2025selfdcreasonactself, qian2025smart}. Concretely, Conversational Agents have been shown to benefit greatly from understanding whether to generate a response directly or invoke specialized tools -- such as APIs, databases, or calculators -- to fill knowledge gaps or execute precise tasks. Specifically, \citet{wang2025selfdcreasonactself} propose Self-DC to adaptively choose between internal reasoning and external acting as needed based on self-aware knowledge boundary, resulting in a better trade-off between effectiveness and efficiency in the context of RAG. SMART~\citep{qian2025smart} further encompasses a wider range of tool use scenarios, with a particular focus on addressing the issue of tool overuse for existing language agents. Subsequently, several studies have followed this direction to enhance metacognition and self-awareness of LLM or agents, with the aim of fostering transparency and responsible AI behavior~\citep{li2025adaptive, qiao2025agentic}.

\noindent\textbf{Self-Correction.} Another essential capability is to learn from mistakes and adapt behaviors in dynamic environments. 
Recent work has explored various approaches to LLM self-refinement, including: allowing models to iteratively improve their own reasoning through feedback~\citep{madaan2024selfrefine}, explicitly verifying intermediate solution steps~\citep{lightman2024lvsbs}, and enabling models to reflect on feedback and store these reflections for continual improvement~\citep{shinn2024reflexion}. RAP~\citep{hao2023rap} conceptualizes reasoning as planning over a learned latent space using a world model, using Monte Carlo Tree Search (MCTS) to explore and select reasoning trajectories. Other approaches such as ETO~\citep{song2024eto} incorporate both successful and failed attempts into training, using reward modeling (e.g., DPO loss) with the corresponding positive and negative trajectories to facilitate more robust refinement. While ETO focuses on overall trajectory outcomes, IPR~\citep{xiong2024ipr} additionally provides intermediate errors to better capture partial failures and reward the decision-making process itself. AgentRefine~\citep{fu2025agentrefine} further integrates multi-turn training data that includes explicit refinement steps following errors, guiding the model back onto a correct path. Finally, Agent-R~\citep{yuan2025agentr} introduces iterative self-training with MCTS-guided critiques, enabling timely self-correction without requiring step-level supervision.

\noindent\textbf{Challenges.} Different LLMs may have different knowledge boundaries and it varies across the time if parameter changes. SFT trained models like SMART~\citep{qian2025smart} offer a streamlined solution but may struggle with generalization across diverse tasks and environments. On the other hand, although self-correction is a highly sought-after capability for intelligent agents, current methods face several obstacles. They often rely on large, carefully curated datasets to teach models how to distinguish between correct and incorrect decisions. Moreover, test-time scaling approaches can be computationally demanding, requiring significant inference time~\citep{zhang2025whatwherehow, li2025system1}. A promising solution would be to reduce the need for such large datasets and enable agents to learn effectively from only a few samples or demonstrations. In the best-case scenario, agents would self-evolve during training by recognizing potential mistakes and correcting them on the fly.


\subsubsection{User \& Interaction Monitoring} 
\label{sec:user-monitoring}

User and interaction monitoring enables Conversational Agents to continuously understand, interpret, and track user behaviors, preferences, and states across interactions, ensuring that agent responses remain relevant, coherent, and contextually appropriate. In this section, we discuss: \textit{User State Tracking}, capturing structured user information and goals; \textit{Personality \& Persona}, facilitating tailored interactions based on user-specific preferences and agent personalities; and \textit{Emotion \& Sentiment}, enhancing conversational effectiveness through empathetic and emotionally attuned responses.

\noindent\textbf{User State Tracking.} State tracking is the mechanism by which a conversational system maintains a persistent and structured memory representation of all relevant user information, goals, and context of interaction over multiple turns~\citep{wang-etal-2021-fast, jacqmin-etal-2022-follow, hu2022icdst, heck2023chatgptzeroshotdst, chung2023instructtods}. Without a coherent record of these evolving attributes—such as preferences, constraints, or previously provided data—an agent is prone to generating inconsistent or irrelevant responses that fail to reflect prior inputs. With the advent of end-to-end systems and LLMs in TOD, recent works have focused on improving state tracking via prompt-based techniques~\citep{feng-etal-2023-towards}, or fine-tuning using specially designed domain-specific conversational datasets like MultiWOZ, SNIPS, or SGD ~\citep{niu2024enhancingdialoguestatetracking}. Comparing user state tracking to API-calling or tool-based language agents highlights its pivotal role in orchestrating external queries and actions. While these newer capabilities enable a model to retrieve real-time information, perform computational tasks, or call APIs as part of the conversation~\citep{li2024fnctod}, coherent state-tracking is what ensures these external operations remain context-appropriate and consistent with user requests. For instance, a restaurant-booking agent may invoke a reservation API only when the dialogue state indicates the user has specified a preferred date, time, and cuisine type. Equally, a research assistant chatbot can reference an academic database through function calls once it confirms the user’s topic, publication year, and format requirements. In both scenarios, state-tracking provides the critical scaffolding for decision-making, enabling the agent to track when, why, and how to invoke external tools—ultimately delivering more accurate, context-sensitive responses that align with the user’s evolving goals~\citep{su-etal-2023-scalable, niu2024enhancingdialoguestatetracking}.

\noindent\textbf{Personality \& Persona.} Personality and persona approaches enable Conversational Agents to exhibit distinct character traits, preferences, and emotional behaviors, making interactions more engaging and human-like. Personalization is built upon retrieving specific portions of memory by tailoring responses to individual user needs, preferences, and past interactions, enhancing trust and user engagement~\citep{salemi2024lamp, generative_agents, chen2024persona}. Recent methods such as BeCand~\citep{lim2023beyond} introduce persona-driven clarification strategies, allowing agents to ask contextually aligned follow-up questions based on distinct persona attributes. Similarly, Personalized Agent Assistants (PAA)~\citep{huang2023personalized} dynamically adapt dialogue strategies and language styles according to learned user preferences, providing tailored conversational experiences. CharacterChat~\citep{tu2023characterchat} incorporates explicit emotional expressions and consistent personality traits by supporting nuanced sentiments and enhancing human-agent social interaction. Finally, OPPU~\citep{tan2024oppu} aims to scale personalization with parameter-efficient fine-tuning (PEFT) by injecting personal PEFT parameters into LLMs, while PER-PCS~\citep{tan2024perpcs} introduces a collaborative framework that shares minimal PEFT pieces, maintaining OPPU-level performance with reduced computation and storage overhead.

\noindent\textbf{Emotion \& Sentiment.} Emotional support is a crucial ability for many conversation scenarios~\citep{ghosal-etal-2020-cosmic, zheng-etal-2023-augesc, deng-etal-2023-knowledge, yan2024talk}, including social interactions, mental health support, and customer service chats. Considering the user's emotions facilitates empathetic conversations, allowing both parties to understand each other's experiences and feelings, which is crucial to establish seamless relationships and is also integral to building trustworthy Conversational Agents. Most of the previous work develops an emotional and empathetic dialogue system in isolation, mainly predict the emotion from a predefined set and generate the corresponding response conditioned on given context and predicted emotion~\citep{zheng-etal-2021-comae,cheng-etal-2023-pal}. Instead, a Conversational Agent seamlessly blends them all into one cohesive conversational flow, regarding user emotion and other statuses as necessary intermediate reasoning steps to reach the final responses~\citep{wang-etal-2023-cue}.

\noindent\textbf{Challenges.} Effective user and interaction monitoring currently faces several concrete challenges. While state tracking aims to maintain context over multiple turns, inherent ambiguities in interpreting user intent and rapidly evolving emotional cues often lead to fragmented or inconsistent representations. Many current approaches rely on superficial methods that detect predefined emotions or static personality traits, failing to capture the nuanced dynamics of user interactions. To address this, we advocate the development of dedicated modules that explicitly capture and update users’ intentions and emotional states. By incorporating ``system 2 thinking'', where the agent engages in deeper, reflective reasoning, the Conversational Agent can generate responses that not only acknowledge but also echo the user’s emotional tone in a contextually appropriate manner~\citep{li2025system1}. Moreover, integrating these modules within a more cohesive digital twin framework~\citep{li2025digitaltwin} can also provide continuous, real-time updates to the agent’s internal state, enabling more robust, proactive, and goal-oriented interactions that enhance trust and overall conversational quality.


\subsection{Control}
\label{sec:control}

Control enables Conversational Agents to perform precise decision-making, ensuring effective tool use and consistent policy adherence. We categorize control into two dimensions: \textit{Tool Utilization} and \textit{Policy Learning \& Following}.

\subsubsection{Tool Utilization} 
\label{sec:tool_utilization}

Tool utilization empowers Conversational Agents to extend their reasoning and interaction beyond internal knowledge by accessing external resources to solve user queries effectively. We focus on two dimensions: \textit{Tool Selection}, identifying the appropriate tool based on context and intent; and \textit{Tool Execution}, deciding when and how to invoke tools.

\noindent\textbf{Tool Selection.}  Selecting the appropriate tool from a set of available options involves identifying the correct function name (e.g., get\_weather(), along with specifying suitable function arguments (e.g., location="Urbana") and argument types (e.g., string). The most straightforward approach involves equipping LLMs with predefined tool calling capabilities~\citep{qin2024toollearningfm, qu2025tool}. Toolformer~\citep{schick2023toolformer} was among the first approaches demonstrating how LLMs can autonomously learn both when and how to invoke external APIs within specific task contexts. 
Following that, Gorilla~\citep{patil2023gorilla} introduced a framework to generate large-scale Python API libraries to facilitate diverse tool usage, while ToolLLM~\citep{qin2024toolllm} expanded the methodology further by providing comprehensive tool integration frameworks coupled with specialized datasets tailored to API usage patterns. 
Granite-Function Calling Model~\citep{abdelaziz2024granite} and xLAM~\citep{zhang2024xlam, zhang2024agentohana} addressed specific challenges such as undefined function calls, incorrect argument types, and argument hallucination by generating a diverse function-calling dataset followed by a post-training stage. Hammer~\citep{lin2025hammer} subsequently extended these developments through an irrelevance-augmented dataset that improves the model's ability to avoid selecting inappropriate functions, combined with a function-masking technique to minimize naming-based misinterpretations and reduce overfitting.  These approaches achieved top performance on the Berkeley Function Calling Leaderboard (BFCL)\footnote{\url{https://gorilla.cs.berkeley.edu/leaderboard.html}} using relatively small-scale, openly available models. 
In parallel to these methods, works such as CREATOR~\citep{qian2023creator} and LATM~\citep{cai2024latm} aim to generate their own tools instead of completely relying on available ones. While these works show promising results in tool selection or creation, challenges remain in determining when and how to execute tools for external information, particularly in multi-turn user interactions.

\noindent\textbf{Tool Execution.} After selecting the appropriate tool, an agent must correctly execute it by interacting with external databases or APIs and retrieving accurate outputs to fulfill user requests~\citep{schick2023toolformer}. Previous approaches such as ToolAlpaca~\citep{tang2023toolalpaca} and ToolLLM~\citep{qin2024toolllm} automatically construct large-scale tool-use corpora and fine-tune LLMs to support generalized and diverse tool usage. Differently, ToolkenGPT~\citep{hao2023toolkengpt} introduces specialized "tool tokens" directly during the language modeling phase, and HuggingGPT~\citep{shen2023hugginggpt} orchestrates external expert models from the HuggingFace library as tools. However, effective tool execution also requires the agent to discern when to perform a function call versus when to respond directly to the user in multi-turn settings. To address these, Agent-Tuning~\citep{zeng2024agenttuning} and FireAct~\citep{chen2023fireact} utilizes SFT on compact yet diverse multi-turn conversational datasets, encompassing both general and tool-oriented agent-specific dialogues. Expanding on this, Agent-FLAN~\citep{chen2024agentflan} further explores optimal dataset design and training methodologies, performing thorough evaluations to validate generalization. Similarly, ToolACE~\citep{liu2025toolace} automates the generation of high-quality and diverse synthetic data, but through a self-evolving pipeline leveraging multi-agent dialogues and dual-layer verification to improve function-execution accuracy. On the other hand, in the TOD domain, FNCTOD~\citep{li2024fnctod} is fine-tuned exclusively on a smaller, domain-specific API dataset tailored specifically for accurate state tracking in limited application contexts. Finally, CoALM~\citep{acikgoz2025singlemodelmastermultiturn} proposes leveraging multi-turn conversational skills and tool-use capabilities through ReAct-style training on multi-task datasets that span both TOD and language agent domains, advancing toward unified Conversational Agents.

\noindent\textbf{Challenges.} Tool utilization in Conversational Agents faces limitations in accurately selecting the appropriate tools and executing them effectively without unnecessary invocations or redundant interactions. Current methods often suffer from issues such as undefined function calls, argument hallucination, incorrect argument type~\citep{zhang2024xlam}, and inefficient tool use~\citep{qian2025smart} due to limited meta-cognitive awareness. RL approaches offer promising solutions for this issue thanks to their customizable and flexible reward mechanisms~\citep{}, but they are under-explored in tool-learning. Further exploration of proactive mechanisms, similar to digital twin concepts, could also yield agents capable of anticipatory and context-sensitive interactions, substantially improving the overall user experience.

\subsubsection{Policy Learning \& Following}
\label{sec:policy_following}

Policy learning and following defined policies are well studied problems in traditional dialogue systems~\citep{young02_icslp, roberto2000, wen-etal-2017-network, bing2017policy, peng-etal-2018-deep, geishauser-etal-2022-dynamic}, but are often overlooked in current systems and agents. User-defined policies are desirable in Conversational Agents to maintain controllability and strict instruction adherence, which otherwise could degrade into inconsistent or even hallucinated behaviors in complex tasks. \cite{hudecek-dusek-2023-are_llms_tod} first highlight that while instruction-tuned LLMs can complete dialogues plausibly, they often fail to track belief states and follow policies without explicit grounding. Building on this insight, SGPTOD~\citep{zhang2023sgptod} proposes schema-guided prompting, where structured policy and belief state information is explicitly injected into LLM inputs to enforce better policy adherence without requiring fine-tuning. FlowBench~\citep{xiao2024flowbench} extends this direction by formalizing dialogue flows as structured workflows and systematically benchmarking how well LLMs align with them, revealing that even with explicit flow guidance, models often deviate under distribution shifts. Extending these insights, AutoTOD~\citep{xu2024autotod} unified the modular TOD components into a single instruction-following model guided by explicit API schemas, allowing agents to autonomously adhere to complex dialogue policies while providing greater controllability. On the other hand, in the $\tau$-Bench~\citep{Yao2024tau} Benchmark, agents are required to follow a predefined dialogue policy during conversations, which makes the task more challenging as they must adhere to these constraints while simultaneously interacting with users to fulfill their intents.

\noindent\textbf{Challenges.} Policy-following remains underexplored yet is critical for effective Conversational Agents, which also relates to long and multi-turn instruction following capabilities of LLMs. As policies grow in length, agents must memorize these extensive instructions while interacting with users. For instance, in $\tau$-Bench, many agents fail to adhere to policies once the length of the conversation increases, leading them to forget or violate policy rules—one of the most frequently reported failure scenarios~\citep{Yao2024tau}. Moreover, providing policies as a single long text may not be the most effective approach for agents to interpret and act upon them. More efficient retrieval-oriented or structured methods could be promising directions for future investigation \citep{xiao2024flowbench}.

\section{Research Roadmap Towards Better Conversational Agents}
\label{sec:future_work}

As the advancements in LLMs, language agents, and conversational systems accelerate, several critical areas emerge as vital avenues for future research on Conversational Agents. This section identifies and elaborates on these directions. 

\noindent\textbf{Long-term Multi-turn Reasoning and Policy Alignment.} One persistent obstacle in designing reliable Conversational Agents is their tendency to lose track of user intent and dialogue context over extended multi-turn interactions~\citep{zhang2025multiturnsurvey}. Existing LLMs often struggle with state tracking when the conversation depth increases. This limitation becomes especially problematic for policy-driven systems (e.g., travel agencies) where failing to recall terms (e.g., ticket cancellation policies) can lead to inaccurate or policy-violating recommendations~\citep{Yao2024tau}. Future research can explore new in-context learning or memory augmentation techniques to reinforce multi-turn context. Dynamic dialogue state updates based on explicit discourse representations look promising, but need stronger defenses against misleading inputs. Maintaining policy alignment across multi-turn conversations also requires continuous safeguards (e.g, multi-step verification or external function calls) though these remain challenging to execute reliably. Together, these strategies can foster more trustworthy, coherent, and policy-compliant agents.

\noindent\textbf{Self-Evolution Capabilities.} Yet another intriguing direction for future research is self-evolving agents that can harness RL to refine their decision-making process online~\citep{fu2025agentrefine, guo2025deepseek}. By generating large-scale interaction trajectories and dynamically incorporating API calls, these models can continuously adjust their reasoning processes without relying on extensive offline fine-tuning. Previous work has shown that LLMs can enhance problem solving abilities by iteratively evaluating and updating their internal reasoning steps (e.g., solving complex math questions through self-refinement)~\citep{guo2025deepseek}, but it remained underexplored in agentic domain. Extending this approach to multi-turn dialogue and tool usage would enable models to better navigate intricate user requests and update dialogue states autonomously. The primary challenge involves ensuring robust reward modeling and preventing pathological self-reinforcement, where unregulated updates could induce undesired behaviors. Successful adoption of such techniques could yield agents that are more adaptable and capable of evolving their skills during real-time interactions.

\noindent\textbf{Evaluating Conversational Agents.} Current methods for evaluating Conversational Agents rely on static offline benchmarks that are susceptible to data contamination, often producing misleading assessments of model capabilities~\citep{sainz2023nlpcontamination, deng2024investigatingcontamination}. Over-reliance on prerecorded dialogue datasets fosters overfitting to specific benchmarks rather than genuine generalization. As a consequence, results obtained from these benchmarks fail to align with actual user experiences in dynamic, interactive environments like in real-world settings. To address this, research can focus on online evaluation frameworks that prevent overfitting and data contamination, using realistic, interactive scenarios (e.g., online reservation from websites) where agents directly engage with dynamic content and complex elements (e.g., changing layouts or pop-ups). Meanwhile, user-centric evaluation metrics should also supplement traditional computational measures. While automated metrics provide quantifiable comparisons, they often fail to capture key aspects of user satisfaction~\citep{liu2016not, ghandeharioun2019approximating, ultes2021blending}. Moreover, future benchmarks can incorporate measures of conversation efficiency (task completion time and effort), user cognitive load, and long-term engagement patterns.

\noindent\textbf{New Learning Methods.} Previous works often train agents by leveraging the latest LLMs and constructing domain-specific datasets for fine-tuning. This process can yield compelling results on specific leaderboards, but some major issues arise: (i) each new base LLM with improved reasoning or knowledge must be repeatedly fine-tuned at high cost, and (ii) specialized fine-tuning often leads to degraded performance in unseen scenarios, indicating weak out-of-distribution generalization. These limitations increase computational and data-collection overhead and undermine the adaptability of deployed agents. A promising alternative involves exploring RL approaches that facilitate online policy updates without the need for extensive SFT. Similar to self-evolution, by continuously learning from real-time interactions and updates its parameters, agents can adapt more efficiently to maintain robust performance across diverse environments.

\noindent\textbf{Collaborative and Multi-Agent Task Completion.} Current Conversational Agents typically operate independently, focusing on single-agent scenarios that limit their effectiveness. Multi-agent coordination and collaboration remains underexplored despite its substantial potential to enhance task efficiency, distribute workloads effectively, and enable agents to jointly handle intricate dialogues requiring varied expertise~\citep{wu2024autogen}. Future research can address inter-agent communication protocols, dynamic role assignments, and the synchronization of shared contexts to facilitate coherent multi-agent dialogues. Leveraging multi-agent reasoning paradigms can enable multiple agents to collaboratively explore and refine diverse reasoning paths, effectively addressing complex tasks where individual agents might fail. Furthermore, establishing robust evaluation frameworks to quantify both individual agent contributions and overall collaborative effectiveness will be essential for advancing multi-agent conversational capabilities~\citep{chen2024llmarena, zhu2025multiagentbench}.

\noindent\textbf{Personalized Conversations.} Current personalization approaches remain superficial, typically limited to static preferences or simple recall of past interactions. Due to the significant variability among user profiles, generalizing personalization to all users is challenging~\citep{chen2024personalizationsurvey}. Future research should focus on techniques that quickly adapt from few samples or demonstrations, enabling agents to dynamically respond to evolving user goals, emotions, and interaction styles, thereby ensuring contextualized and effective interactions.

\noindent\textbf{Proactivity.} A particularly underdeveloped frontier is proactivity: most agents today are fundamentally reactive, responding only to explicit user prompts~\citep{lu2024proactive}. In contrast, proactive Conversational Agents can anticipate needs, take initiative, and structure conversations. These agents must plan conversational trajectories, evaluate possible interaction outcomes, and decide when and how to intervene or steer a dialogue. Thus, planning is essential for proactivity: to act effectively, proactive agents must forecast the impact of their actions not only on task success but also on the user’s preferences, mental state, and future behavior. Unlike reactive agents that operate turn by turn, proactive agents require dialogue-level foresight, balancing initiative-taking with adaptability and user trust. Future agents should both interpret these features in user input and incorporate them into their own responses.

\noindent\textbf{Multimodal Conversational Agents.}
As Conversational Agents evolve beyond text-only paradigms~\citep{xie2024largemultimodalagents, ma2024surveyembodied}, developing robust multimodal capabilities emerges as a critical frontier for future research~\citep{xi2025risesurveyagents, liu2025acognitive}. Current agents primarily excel in linguistic processing, yet human communication inherently spans multiple sensory channels simultaneously, combining speech, vision, and gesture. The prosodic elements of human speech (e.g., intonation, rhythm, and stress) carry crucial information often lost in text transcriptions. Future agents should both interpret these features in user input and incorporate them into their own responses.

\vspace{-2mm}
\section{Final Remarks}
\vspace{-2mm}

Our desideratum introduces a structured definition of Conversational Agents, emphasizing their essential capabilities, highlighting current limitations, and identifying emerging capabilities needed for further advancement. Our motivation for categorizing Conversational Agents into reasoning, monitoring, and control dimensions is to provide clarity and structured guidance for ongoing and future research. We believe that the potential of Conversational Agents to significantly progress towards AGI is substantial. Through our work, we hope to encourage deeper discussions and foster research developments, particularly focusing on (i) new and realistic benchmarks, (ii) multi-turn reasoning and long-term policy following, (iii) cultivating self-evolution capabilities, (iv) enabling deeper personalization, and (v) fostering more collaborative proactive engagements with users.

\section{Limitations: What This Work Does Not Cover}

Although this work provides a comprehensive overview of Conversational Agents based on our proposed taxonomy, there are several important aspects that remain outside the scope of this work. Memory, a crucial cognitive component of agents, intersects with multiple elements of our framework including reasoning (short-term memory), user state tracking (both short and long-term memory), and personalization (short and long-term memory). We acknowledge the significance of memory systems, which have been thoroughly examined in previous surveys~\citep{sumers2024cognitive, wang2024survey, xi2025risesurveyagents}. Similarly, we do not dedicate a separate section to planning, even though it intersects with reasoning and is briefly mentioned in that context. Additionally, our desiderata intentionally focus on text-only conversational agents, excluding multimodal capabilities. This boundary allows us to address fundamental challenges in the text domain before extending to additional modalities. We also acknowledge that safety considerations for agents while interacting with users are critically important~\citep{liu2025acognitive}, but not covered in this work. We hope that our proposed desiderata and the listed resources can serve as a useful foundation for researchers aiming to build more capable and aligned Conversational Agents.

\section*{Acknowledgments}
We would like to express our sincere gratitude to our colleagues at UIUC (Ishika Agarwal, Abhinav Chinta, Beyza Bozdag, and Suvodip Dey) for their discussions and constructive feedback on this work. We also acknowledge that our taxonomy is adapted and modified from \cite{mondorf2024beyond}.

\bibliography{colm2025_conference}

\begin{thebibliography}{151}
\providecommand{\natexlab}[1]{#1}
\providecommand{\url}[1]{\texttt{#1}}
\expandafter\ifx\csname urlstyle\endcsname\relax
  \providecommand{\doi}[1]{doi: #1}\else
  \providecommand{\doi}{doi: \begingroup \urlstyle{rm}\Url}\fi

\bibitem[Abdelaziz et~al.(2024)Abdelaziz, Basu, Agarwal, Kumaravel, Stallone, Panda, Rizk, Bhargav, Crouse, Gunasekara, Ikbal, Joshi, Karanam, Kumar, Munawar, Neelam, Raghu, Sharma, Soria, Sreedhar, Venkateswaran, Unuvar, Cox, Roukos, Lastras, and Kapanipathi]{abdelaziz2024granite}
Ibrahim Abdelaziz, Kinjal Basu, Mayank Agarwal, Sadhana Kumaravel, Matthew Stallone, Rameswar Panda, Yara Rizk, G~P~Shrivatsa Bhargav, Maxwell Crouse, Chulaka Gunasekara, Shajith Ikbal, Sachindra Joshi, Hima Karanam, Vineet Kumar, Asim Munawar, Sumit Neelam, Dinesh Raghu, Udit Sharma, Adriana~Meza Soria, Dheeraj Sreedhar, Praveen Venkateswaran, Merve Unuvar, David~Daniel Cox, Salim Roukos, Luis~A. Lastras, and Pavan Kapanipathi.
\newblock Granite-function calling model: Introducing function calling abilities via multi-task learning of granular tasks.
\newblock In Franck Dernoncourt, Daniel Preo{\c{t}}iuc-Pietro, and Anastasia Shimorina (eds.), \emph{Proceedings of the 2024 Conference on Empirical Methods in Natural Language Processing: Industry Track}, pp.\  1131--1139, Miami, Florida, US, November 2024. Association for Computational Linguistics.
\newblock \doi{10.18653/v1/2024.emnlp-industry.85}.
\newblock URL \url{https://aclanthology.org/2024.emnlp-industry.85/}.

\bibitem[Achiam et~al.(2023)]{Achiam2023GPT4TR}
OpenAI~Josh Achiam et~al.
\newblock Gpt-4 technical report.
\newblock \emph{arXiv preprint arXiv:2303.08774}, 2023.
\newblock URL \url{https://api.semanticscholar.org/CorpusID:257532815}.

\bibitem[Acikgoz et~al.(2025)Acikgoz, Greer, Datta, Yang, Zeng, Elachqar, Koukoumidis, Hakkani-Tür, and Tur]{acikgoz2025singlemodelmastermultiturn}
Emre~Can Acikgoz, Jeremiah Greer, Akul Datta, Ze~Yang, William Zeng, Oussama Elachqar, Emmanouil Koukoumidis, Dilek Hakkani-Tür, and Gokhan Tur.
\newblock Can a single model master both multi-turn conversations and tool use? coalm: A unified conversational agentic language model, 2025.
\newblock URL \url{https://arxiv.org/abs/2502.08820}.

\bibitem[Amayuelas et~al.(2024{\natexlab{a}})Amayuelas, Wong, Pan, Chen, and Wang]{amayuelas-etal-2024-knowledge}
Alfonso Amayuelas, Kyle Wong, Liangming Pan, Wenhu Chen, and William~Yang Wang.
\newblock Knowledge of knowledge: Exploring known-unknowns uncertainty with large language models.
\newblock In Lun-Wei Ku, Andre Martins, and Vivek Srikumar (eds.), \emph{Findings of the Association for Computational Linguistics: ACL 2024}, pp.\  6416--6432, Bangkok, Thailand, August 2024{\natexlab{a}}. Association for Computational Linguistics.
\newblock \doi{10.18653/v1/2024.findings-acl.383}.
\newblock URL \url{https://aclanthology.org/2024.findings-acl.383/}.

\bibitem[Amayuelas et~al.(2024{\natexlab{b}})Amayuelas, Wong, Pan, Chen, and Wang]{amayuelas2024kuq}
Alfonso Amayuelas, Kyle Wong, Liangming Pan, Wenhu Chen, and William~Yang Wang.
\newblock Knowledge of knowledge: Exploring known-unknowns uncertainty with large language models.
\newblock In \emph{Findings of the Association for Computational Linguistics ACL 2024}, pp.\  6416--6432, 2024{\natexlab{b}}.

\bibitem[Andukuri et~al.(2024)Andukuri, Fr{\"a}nken, Gerstenberg, and Goodman]{andukuri2024star}
Chinmaya Andukuri, Jan-Philipp Fr{\"a}nken, Tobias Gerstenberg, and Noah Goodman.
\newblock Star-gate: Teaching language models to ask clarifying questions.
\newblock In \emph{First Conference on Language Modeling}, 2024.

\bibitem[Cai et~al.(2024)Cai, Wang, Ma, Chen, and Zhou]{cai2024latm}
Tianle Cai, Xuezhi Wang, Tengyu Ma, Xinyun Chen, and Denny Zhou.
\newblock Large language models as tool makers.
\newblock In \emph{The Twelfth International Conference on Learning Representations}, 2024.
\newblock URL \url{https://openreview.net/forum?id=qV83K9d5WB}.

\bibitem[Chen et~al.(2023{\natexlab{a}})Chen, Shu, Shareghi, Collier, Narasimhan, and Yao]{chen2023fireact}
Baian Chen, Chang Shu, Ehsan Shareghi, Nigel Collier, Karthik Narasimhan, and Shunyu Yao.
\newblock Fireact: Toward language agent fine-tuning.
\newblock \emph{arXiv preprint arXiv:2310.05915}, 2023{\natexlab{a}}.

\bibitem[Chen et~al.(2017)Chen, Liu, Yin, and Tang]{chends_survey}
Hongshen Chen, Xiaorui Liu, Dawei Yin, and Jiliang Tang.
\newblock A survey on dialogue systems: Recent advances and new frontiers.
\newblock \emph{SIGKDD Explor. Newsl.}, 19\penalty0 (2):\penalty0 25–35, November 2017.
\newblock ISSN 1931-0145.
\newblock \doi{10.1145/3166054.3166058}.
\newblock URL \url{https://doi.org/10.1145/3166054.3166058}.

\bibitem[Chen et~al.(2024{\natexlab{a}})Chen, Wang, Xu, Yuan, Zhang, Shi, Xie, Li, Yang, Zhu, et~al.]{chen2024persona}
Jiangjie Chen, Xintao Wang, Rui Xu, Siyu Yuan, Yikai Zhang, Wei Shi, Jian Xie, Shuang Li, Ruihan Yang, Tinghui Zhu, et~al.
\newblock From persona to personalization: A survey on role-playing language agents.
\newblock \emph{arXiv preprint arXiv:2404.18231}, 2024{\natexlab{a}}.

\bibitem[Chen et~al.(2024{\natexlab{b}})Chen, Liu, Huang, Wu, Liu, Jiang, Pu, Lei, Chen, Wang, et~al.]{chen2024personalizationsurvey}
Jin Chen, Zheng Liu, Xu~Huang, Chenwang Wu, Qi~Liu, Gangwei Jiang, Yuanhao Pu, Yuxuan Lei, Xiaolong Chen, Xingmei Wang, et~al.
\newblock When large language models meet personalization: Perspectives of challenges and opportunities.
\newblock \emph{World Wide Web}, 27\penalty0 (4):\penalty0 42, 2024{\natexlab{b}}.

\bibitem[Chen et~al.(2024{\natexlab{c}})Chen, Hu, Liu, Huang, Tu, He, and Wen]{chen2024llmarena}
Junzhe Chen, Xuming Hu, Shuodi Liu, Shiyu Huang, Wei-Wei Tu, Zhaofeng He, and Lijie Wen.
\newblock {LLMA}rena: Assessing capabilities of large language models in dynamic multi-agent environments.
\newblock In Lun-Wei Ku, Andre Martins, and Vivek Srikumar (eds.), \emph{Proceedings of the 62nd Annual Meeting of the Association for Computational Linguistics (Volume 1: Long Papers)}, pp.\  13055--13077, Bangkok, Thailand, August 2024{\natexlab{c}}. Association for Computational Linguistics.
\newblock \doi{10.18653/v1/2024.acl-long.705}.
\newblock URL \url{https://aclanthology.org/2024.acl-long.705/}.

\bibitem[Chen et~al.(2023{\natexlab{b}})Chen, Ma, Wang, and Cohen]{chen2023pot}
Wenhu Chen, Xueguang Ma, Xinyi Wang, and William~W. Cohen.
\newblock Program of thoughts prompting: Disentangling computation from reasoning for numerical reasoning tasks.
\newblock \emph{Transactions on Machine Learning Research}, 2023{\natexlab{b}}.
\newblock ISSN 2835-8856.
\newblock URL \url{https://openreview.net/forum?id=YfZ4ZPt8zd}.

\bibitem[Chen et~al.(2024{\natexlab{d}})Chen, Liu, Wang, Zhang, Liu, Lin, Chen, and Zhao]{chen2024agentflan}
Zehui Chen, Kuikun Liu, Qiuchen Wang, Wenwei Zhang, Jiangning Liu, Dahua Lin, Kai Chen, and Feng Zhao.
\newblock Agent-flan: Designing data and methods of effective agent tuning for large language models.
\newblock In \emph{Findings of the Association for Computational Linguistics ACL 2024}, pp.\  9354--9366, 2024{\natexlab{d}}.

\bibitem[Chen et~al.(2023{\natexlab{c}})Chen, Zhou, Zhang, Gong, Zhao, and Wen]{chen2023chatcot}
Zhipeng Chen, Kun Zhou, Beichen Zhang, Zheng Gong, Wayne~Xin Zhao, and Ji-Rong Wen.
\newblock Chatcot: Tool-augmented chain-of-thought reasoning on chat-based large language models.
\newblock \emph{arXiv preprint arXiv:2305.14323}, 2023{\natexlab{c}}.

\bibitem[Cheng et~al.(2023)Cheng, Sabour, Sun, Chen, and Huang]{cheng-etal-2023-pal}
Jiale Cheng, Sahand Sabour, Hao Sun, Zhuang Chen, and Minlie Huang.
\newblock {PAL}: Persona-augmented emotional support conversation generation.
\newblock In Anna Rogers, Jordan Boyd-Graber, and Naoaki Okazaki (eds.), \emph{Findings of the Association for Computational Linguistics: ACL 2023}, pp.\  535--554, Toronto, Canada, July 2023. Association for Computational Linguistics.
\newblock \doi{10.18653/v1/2023.findings-acl.34}.
\newblock URL \url{https://aclanthology.org/2023.findings-acl.34/}.

\bibitem[Cheng et~al.(2024)Cheng, Sun, Liu, Zhang, Yin, Li, Li, He, Chen, and Qiu]{cheng2024idk}
Qinyuan Cheng, Tianxiang Sun, Xiangyang Liu, Wenwei Zhang, Zhangyue Yin, Shimin Li, Linyang Li, Zhengfu He, Kai Chen, and Xipeng Qiu.
\newblock Can ai assistants know what they don’t know?
\newblock In \emph{International Conference on Machine Learning}, pp.\  8184--8202. PMLR, 2024.

\bibitem[Chung et~al.(2023{\natexlab{a}})Chung, Cahyawijaya, Wilie, Lovenia, and Fung]{chung-etal-2023-instructtods}
Willy Chung, Samuel Cahyawijaya, Bryan Wilie, Holy Lovenia, and Pascale Fung.
\newblock {I}nstruct{TODS}: Large language models for end-to-end task-oriented dialogue systems.
\newblock In Kehai Chen and Lun-Wei Ku (eds.), \emph{Proceedings of the Second Workshop on Natural Language Interfaces}, pp.\  1--21, Bali, Indonesia, November 2023{\natexlab{a}}. Association for Computational Linguistics.
\newblock \doi{10.18653/v1/2023.nlint-1.1}.
\newblock URL \url{https://aclanthology.org/2023.nlint-1.1/}.

\bibitem[Chung et~al.(2023{\natexlab{b}})Chung, Cahyawijaya, Wilie, Lovenia, and Fung]{chung2023instructtods}
Willy Chung, Samuel Cahyawijaya, Bryan Wilie, Holy Lovenia, and Pascale Fung.
\newblock {I}nstruct{TODS}: Large language models for end-to-end task-oriented dialogue systems.
\newblock In Kehai Chen and Lun-Wei Ku (eds.), \emph{Proceedings of the Second Workshop on Natural Language Interfaces}, pp.\  1--21, Bali, Indonesia, November 2023{\natexlab{b}}. Association for Computational Linguistics.
\newblock \doi{10.18653/v1/2023.nlint-1.1}.
\newblock URL \url{https://aclanthology.org/2023.nlint-1.1/}.

\bibitem[Deng et~al.(2024)Deng, Zhao, Tang, Gerstein, and Cohan]{deng2024investigatingcontamination}
Chunyuan Deng, Yilun Zhao, Xiangru Tang, Mark Gerstein, and Arman Cohan.
\newblock Investigating data contamination in modern benchmarks for large language models.
\newblock In Kevin Duh, Helena Gomez, and Steven Bethard (eds.), \emph{Proceedings of the 2024 Conference of the North American Chapter of the Association for Computational Linguistics: Human Language Technologies (Volume 1: Long Papers)}, pp.\  8706--8719, Mexico City, Mexico, June 2024. Association for Computational Linguistics.
\newblock \doi{10.18653/v1/2024.naacl-long.482}.
\newblock URL \url{https://aclanthology.org/2024.naacl-long.482/}.

\bibitem[Deng et~al.(2023)Deng, Zhang, Yuan, and Lam]{deng-etal-2023-knowledge}
Yang Deng, Wenxuan Zhang, Yifei Yuan, and Wai Lam.
\newblock Knowledge-enhanced mixed-initiative dialogue system for emotional support conversations.
\newblock In Anna Rogers, Jordan Boyd-Graber, and Naoaki Okazaki (eds.), \emph{Proceedings of the 61st Annual Meeting of the Association for Computational Linguistics (Volume 1: Long Papers)}, pp.\  4079--4095, Toronto, Canada, July 2023. Association for Computational Linguistics.
\newblock \doi{10.18653/v1/2023.acl-long.225}.
\newblock URL \url{https://aclanthology.org/2023.acl-long.225/}.

\bibitem[Dongre et~al.(2024)Dongre, Yang, Acikgoz, Dey, Tur, and Hakkani-T{\"u}r]{dongre2024respact}
Vardhan Dongre, Xiaocheng Yang, Emre~Can Acikgoz, Suvodip Dey, Gokhan Tur, and Dilek Hakkani-T{\"u}r.
\newblock Respact: Harmonizing reasoning, speaking, and acting towards building large language model-based conversational ai agents.
\newblock \emph{arXiv preprint arXiv:2411.00927}, 2024.

\bibitem[Dubey et~al.(2024)]{Dubey2024TheL3-llama3}
Abhimanyu Dubey et~al.
\newblock The llama 3 herd of models.
\newblock \emph{ArXiv}, abs/2407.21783, 2024.
\newblock URL \url{https://api.semanticscholar.org/CorpusID:271571434}.

\bibitem[Feng et~al.(2023{\natexlab{a}})Feng, Lu, Liu, Zhan, and Wu]{feng-etal-2023-ldst}
Yujie Feng, Zexin Lu, Bo~Liu, Liming Zhan, and Xiao-Ming Wu.
\newblock Towards {LLM}-driven dialogue state tracking.
\newblock In Houda Bouamor, Juan Pino, and Kalika Bali (eds.), \emph{Proceedings of the 2023 Conference on Empirical Methods in Natural Language Processing}, pp.\  739--755, Singapore, December 2023{\natexlab{a}}. Association for Computational Linguistics.
\newblock \doi{10.18653/v1/2023.emnlp-main.48}.
\newblock URL \url{https://aclanthology.org/2023.emnlp-main.48/}.

\bibitem[Feng et~al.(2023{\natexlab{b}})Feng, Lu, Liu, Zhan, and Wu]{feng-etal-2023-towards}
Yujie Feng, Zexin Lu, Bo~Liu, Liming Zhan, and Xiao-Ming Wu.
\newblock Towards {LLM}-driven dialogue state tracking.
\newblock In Houda Bouamor, Juan Pino, and Kalika Bali (eds.), \emph{Proceedings of the 2023 Conference on Empirical Methods in Natural Language Processing}, pp.\  739--755, Singapore, December 2023{\natexlab{b}}. Association for Computational Linguistics.
\newblock \doi{10.18653/v1/2023.emnlp-main.48}.
\newblock URL \url{https://aclanthology.org/2023.emnlp-main.48/}.

\bibitem[Feng et~al.(2023{\natexlab{c}})Feng, Lu, Liu, Zhan, and Wu]{feng2023ldst}
Yujie Feng, Zexin Lu, Bo~Liu, Liming Zhan, and Xiao-Ming Wu.
\newblock Towards {LLM}-driven dialogue state tracking.
\newblock In Houda Bouamor, Juan Pino, and Kalika Bali (eds.), \emph{Proceedings of the 2023 Conference on Empirical Methods in Natural Language Processing}, pp.\  739--755, Singapore, December 2023{\natexlab{c}}. Association for Computational Linguistics.
\newblock \doi{10.18653/v1/2023.emnlp-main.48}.
\newblock URL \url{https://aclanthology.org/2023.emnlp-main.48/}.

\bibitem[Fu et~al.(2025)Fu, He, Wang, Hong, GongQue, Zeng, Wang, Wang, Cai, and Xu]{fu2025agentrefine}
Dayuan Fu, Keqing He, Yejie Wang, Wentao Hong, Zhuoma GongQue, Weihao Zeng, Wei Wang, Jingang Wang, Xunliang Cai, and Weiran Xu.
\newblock Agentrefine: Enhancing agent generalization through refinement tuning.
\newblock In \emph{The Thirteenth International Conference on Learning Representations}, 2025.
\newblock URL \url{https://openreview.net/forum?id=FDimWzmcWn}.

\bibitem[Fu et~al.(2023)Fu, Inoue, Chu, and Kawahara]{fu-etal-2023-reasoning}
Yahui Fu, Koji Inoue, Chenhui Chu, and Tatsuya Kawahara.
\newblock Reasoning before responding: Integrating commonsense-based causality explanation for empathetic response generation.
\newblock In Svetlana Stoyanchev, Shafiq Joty, David Schlangen, Ondrej Dusek, Casey Kennington, and Malihe Alikhani (eds.), \emph{Proceedings of the 24th Annual Meeting of the Special Interest Group on Discourse and Dialogue}, pp.\  645--656, Prague, Czechia, September 2023. Association for Computational Linguistics.
\newblock \doi{10.18653/v1/2023.sigdial-1.60}.
\newblock URL \url{https://aclanthology.org/2023.sigdial-1.60/}.

\bibitem[Gao et~al.(2023)Gao, Madaan, Zhou, Alon, Liu, Yang, Callan, and Neubig]{gao2023pal}
Luyu Gao, Aman Madaan, Shuyan Zhou, Uri Alon, Pengfei Liu, Yiming Yang, Jamie Callan, and Graham Neubig.
\newblock {PAL}: Program-aided language models.
\newblock In Andreas Krause, Emma Brunskill, Kyunghyun Cho, Barbara Engelhardt, Sivan Sabato, and Jonathan Scarlett (eds.), \emph{Proceedings of the 40th International Conference on Machine Learning}, volume 202 of \emph{Proceedings of Machine Learning Research}, pp.\  10764--10799. PMLR, 23--29 Jul 2023.
\newblock URL \url{https://proceedings.mlr.press/v202/gao23f.html}.

\bibitem[Geishauser et~al.(2022)Geishauser, van Niekerk, Lin, Lubis, Heck, Feng, and Ga{\v{s}}i{\'c}]{geishauser-etal-2022-dynamic}
Christian Geishauser, Carel van Niekerk, Hsien-chin Lin, Nurul Lubis, Michael Heck, Shutong Feng, and Milica Ga{\v{s}}i{\'c}.
\newblock Dynamic dialogue policy for continual reinforcement learning.
\newblock In Nicoletta Calzolari, Chu-Ren Huang, Hansaem Kim, James Pustejovsky, Leo Wanner, Key-Sun Choi, Pum-Mo Ryu, Hsin-Hsi Chen, Lucia Donatelli, Heng Ji, Sadao Kurohashi, Patrizia Paggio, Nianwen Xue, Seokhwan Kim, Younggyun Hahm, Zhong He, Tony~Kyungil Lee, Enrico Santus, Francis Bond, and Seung-Hoon Na (eds.), \emph{Proceedings of the 29th International Conference on Computational Linguistics}, pp.\  266--284, Gyeongju, Republic of Korea, October 2022. International Committee on Computational Linguistics.
\newblock URL \url{https://aclanthology.org/2022.coling-1.21/}.

\bibitem[Ghandeharioun et~al.(2019)Ghandeharioun, Shen, Jaques, Ferguson, Jones, Lapedriza, and Picard]{ghandeharioun2019approximating}
Asma Ghandeharioun, Judy~Hanwen Shen, Natasha Jaques, Craig Ferguson, Noah Jones, Agata Lapedriza, and Rosalind Picard.
\newblock Approximating interactive human evaluation with self-play for open-domain dialog systems.
\newblock \emph{Advances in Neural Information Processing Systems}, 32, 2019.

\bibitem[Ghosal et~al.(2020)Ghosal, Majumder, Gelbukh, Mihalcea, and Poria]{ghosal-etal-2020-cosmic}
Deepanway Ghosal, Navonil Majumder, Alexander Gelbukh, Rada Mihalcea, and Soujanya Poria.
\newblock {COSMIC}: {CO}mmon{S}ense knowledge for e{M}otion identification in conversations.
\newblock In Trevor Cohn, Yulan He, and Yang Liu (eds.), \emph{Findings of the Association for Computational Linguistics: EMNLP 2020}, pp.\  2470--2481, Online, November 2020. Association for Computational Linguistics.
\newblock \doi{10.18653/v1/2020.findings-emnlp.224}.
\newblock URL \url{https://aclanthology.org/2020.findings-emnlp.224/}.

\bibitem[Guo et~al.(2025)Guo, Yang, Zhang, Song, Zhang, Xu, Zhu, Ma, Wang, Bi, et~al.]{guo2025deepseek}
Daya Guo, Dejian Yang, Haowei Zhang, Junxiao Song, Ruoyu Zhang, Runxin Xu, Qihao Zhu, Shirong Ma, Peiyi Wang, Xiao Bi, et~al.
\newblock Deepseek-r1: Incentivizing reasoning capability in llms via reinforcement learning.
\newblock \emph{arXiv preprint arXiv:2501.12948}, 2025.

\bibitem[Hao et~al.(2023{\natexlab{a}})Hao, Gu, Ma, Hong, Wang, Wang, and Hu]{hao2023rap}
Shibo Hao, Yi~Gu, Haodi Ma, Joshua Hong, Zhen Wang, Daisy Wang, and Zhiting Hu.
\newblock Reasoning with language model is planning with world model.
\newblock In \emph{Proceedings of the 2023 Conference on Empirical Methods in Natural Language Processing}, pp.\  8154--8173, 2023{\natexlab{a}}.

\bibitem[Hao et~al.(2023{\natexlab{b}})Hao, Gu, Ma, Hong, Wang, Wang, and Hu]{hao2023reasoning}
Shibo Hao, Yi~Gu, Haodi Ma, Joshua Hong, Zhen Wang, Daisy Wang, and Zhiting Hu.
\newblock Reasoning with language model is planning with world model.
\newblock In \emph{Proceedings of the 2023 Conference on Empirical Methods in Natural Language Processing}, pp.\  8154--8173, 2023{\natexlab{b}}.

\bibitem[Hao et~al.(2023{\natexlab{c}})Hao, Liu, Wang, and Hu]{hao2023toolkengpt}
Shibo Hao, Tianyang Liu, Zhen Wang, and Zhiting Hu.
\newblock Toolkengpt: Augmenting frozen language models with massive tools via tool embeddings.
\newblock \emph{Advances in neural information processing systems}, 36:\penalty0 45870--45894, 2023{\natexlab{c}}.

\bibitem[Heck et~al.(2023)Heck, Lubis, Ruppik, Vukovic, Feng, Geishauser, Lin, van Niekerk, and Gasic]{heck2023chatgptzeroshotdst}
Michael Heck, Nurul Lubis, Benjamin Ruppik, Renato Vukovic, Shutong Feng, Christian Geishauser, Hsien-chin Lin, Carel van Niekerk, and Milica Gasic.
\newblock {C}hat{GPT} for zero-shot dialogue state tracking: A solution or an opportunity?
\newblock In Anna Rogers, Jordan Boyd-Graber, and Naoaki Okazaki (eds.), \emph{Proceedings of the 61st Annual Meeting of the Association for Computational Linguistics (Volume 2: Short Papers)}, pp.\  936--950, Toronto, Canada, July 2023. Association for Computational Linguistics.
\newblock \doi{10.18653/v1/2023.acl-short.81}.
\newblock URL \url{https://aclanthology.org/2023.acl-short.81/}.

\bibitem[Hu et~al.(2022)Hu, Lee, Xie, Yu, Smith, and Ostendorf]{hu2022icdst}
Yushi Hu, Chia-Hsuan Lee, Tianbao Xie, Tao Yu, Noah~A. Smith, and Mari Ostendorf.
\newblock In-context learning for few-shot dialogue state tracking.
\newblock In Yoav Goldberg, Zornitsa Kozareva, and Yue Zhang (eds.), \emph{Findings of the Association for Computational Linguistics: EMNLP 2022}, pp.\  2627--2643, Abu Dhabi, United Arab Emirates, December 2022. Association for Computational Linguistics.
\newblock \doi{10.18653/v1/2022.findings-emnlp.193}.
\newblock URL \url{https://aclanthology.org/2022.findings-emnlp.193/}.

\bibitem[Huang et~al.(2023)Huang, Zhang, Ko, Liu, Wu, Wang, and Tang]{huang2023personalized}
Qiushi Huang, Yu~Zhang, Tom Ko, Xubo Liu, Bo~Wu, Wenwu Wang, and H~Tang.
\newblock Personalized dialogue generation with persona-adaptive attention.
\newblock In \emph{Proceedings of the AAAI Conference on Artificial Intelligence}, volume~37, pp.\  12916--12923, 2023.

\bibitem[Hude{\v{c}}ek \& Dusek(2023)Hude{\v{c}}ek and Dusek]{hudecek-dusek-2023-are_llms_tod}
Vojt{\v{e}}ch Hude{\v{c}}ek and Ondrej Dusek.
\newblock Are large language models all you need for task-oriented dialogue?
\newblock In Svetlana Stoyanchev, Shafiq Joty, David Schlangen, Ondrej Dusek, Casey Kennington, and Malihe Alikhani (eds.), \emph{Proceedings of the 24th Annual Meeting of the Special Interest Group on Discourse and Dialogue}, pp.\  216--228, Prague, Czechia, September 2023. Association for Computational Linguistics.
\newblock \doi{10.18653/v1/2023.sigdial-1.21}.
\newblock URL \url{https://aclanthology.org/2023.sigdial-1.21/}.

\bibitem[Inaba et~al.(2023)Inaba, Kiyomaru, Cheng, and Kurohashi]{inaba2023multitool}
Tatsuro Inaba, Hirokazu Kiyomaru, Fei Cheng, and Sadao Kurohashi.
\newblock Multitool-cot: Gpt-3 can use multiple external tools with chain of thought prompting.
\newblock \emph{arXiv preprint arXiv:2305.16896}, 2023.

\bibitem[Jacqmin et~al.(2022)Jacqmin, Rojas~Barahona, and Favre]{jacqmin-etal-2022-follow}
L{\'e}o Jacqmin, Lina~M. Rojas~Barahona, and Benoit Favre.
\newblock {\textquotedblleft}do you follow me?{\textquotedblright}: A survey of recent approaches in dialogue state tracking.
\newblock In Oliver Lemon, Dilek Hakkani-Tur, Junyi~Jessy Li, Arash Ashrafzadeh, Daniel~Hern{\'a}ndez Garcia, Malihe Alikhani, David Vandyke, and Ond{\v{r}}ej Du{\v{s}}ek (eds.), \emph{Proceedings of the 23rd Annual Meeting of the Special Interest Group on Discourse and Dialogue}, pp.\  336--350, Edinburgh, UK, September 2022. Association for Computational Linguistics.
\newblock \doi{10.18653/v1/2022.sigdial-1.33}.
\newblock URL \url{https://aclanthology.org/2022.sigdial-1.33/}.

\bibitem[Jaech et~al.(2024)Jaech, Kalai, Lerer, Richardson, El-Kishky, Low, Helyar, Madry, Beutel, Carney, et~al.]{jaech2024o1}
Aaron Jaech, Adam Kalai, Adam Lerer, Adam Richardson, Ahmed El-Kishky, Aiden Low, Alec Helyar, Aleksander Madry, Alex Beutel, Alex Carney, et~al.
\newblock Openai o1 system card.
\newblock \emph{arXiv preprint arXiv:2412.16720}, 2024.

\bibitem[Khattab et~al.(2023)Khattab, Singhvi, Maheshwari, Zhang, Santhanam, Vardhamanan, Haq, Sharma, Joshi, Moazam, et~al.]{khattab2023dspy}
Omar Khattab, Arnav Singhvi, Paridhi Maheshwari, Zhiyuan Zhang, Keshav Santhanam, Sri Vardhamanan, Saiful Haq, Ashutosh Sharma, Thomas~T Joshi, Hanna Moazam, et~al.
\newblock Dspy: Compiling declarative language model calls into self-improving pipelines.
\newblock \emph{CoRR}, 2023.

\bibitem[King \& Flanigan(2024)King and Flanigan]{king-flanigan-2024-unsupervised}
Brendan King and Jeffrey Flanigan.
\newblock Unsupervised end-to-end task-oriented dialogue with {LLM}s: The power of the noisy channel.
\newblock In Yaser Al-Onaizan, Mohit Bansal, and Yun-Nung Chen (eds.), \emph{Proceedings of the 2024 Conference on Empirical Methods in Natural Language Processing}, pp.\  8283--8300, Miami, Florida, USA, November 2024. Association for Computational Linguistics.
\newblock \doi{10.18653/v1/2024.emnlp-main.473}.
\newblock URL \url{https://aclanthology.org/2024.emnlp-main.473/}.

\bibitem[Kojima et~al.(2022)Kojima, Gu, Reid, Matsuo, and Iwasawa]{kojima2022zerocot}
Takeshi Kojima, Shixiang~Shane Gu, Machel Reid, Yutaka Matsuo, and Yusuke Iwasawa.
\newblock Large language models are zero-shot reasoners.
\newblock \emph{Advances in neural information processing systems}, 35:\penalty0 22199--22213, 2022.

\bibitem[Kumar et~al.(2025)Kumar, Ashraf, Thawakar, Anwer, Cholakkal, Shah, Yang, Torr, Khan, and Khan]{kumar2025reasoningsurvey}
Komal Kumar, Tajamul Ashraf, Omkar Thawakar, Rao~Muhammad Anwer, Hisham Cholakkal, Mubarak Shah, Ming-Hsuan Yang, Phillip~HS Torr, Salman Khan, and Fahad~Shahbaz Khan.
\newblock Llm post-training: A deep dive into reasoning large language models.
\newblock \emph{arXiv preprint arXiv:2502.21321}, 2025.

\bibitem[Levin et~al.(2000)Levin, Pieraccini, and Eckert]{roberto2000}
E.~Levin, R.~Pieraccini, and W.~Eckert.
\newblock A stochastic model of human-machine interaction for learning dialog strategies.
\newblock \emph{IEEE Transactions on Speech and Audio Processing}, 8\penalty0 (1):\penalty0 11--23, 2000.
\newblock \doi{10.1109/89.817450}.

\bibitem[Li et~al.(2023)Li, Zhao, Yu, Song, Li, Yu, Li, Huang, and Li]{li-etal-2023-api}
Minghao Li, Yingxiu Zhao, Bowen Yu, Feifan Song, Hangyu Li, Haiyang Yu, Zhoujun Li, Fei Huang, and Yongbin Li.
\newblock {API}-bank: A comprehensive benchmark for tool-augmented {LLM}s.
\newblock In Houda Bouamor, Juan Pino, and Kalika Bali (eds.), \emph{Proceedings of the 2023 Conference on Empirical Methods in Natural Language Processing}, pp.\  3102--3116, Singapore, December 2023. Association for Computational Linguistics.
\newblock \doi{10.18653/v1/2023.emnlp-main.187}.
\newblock URL \url{https://aclanthology.org/2023.emnlp-main.187}.

\bibitem[Li et~al.(2024{\natexlab{a}})Li, Zhao, Deng, Zhang, Li, Xie, Ng, and Chua]{li2024knowledgeboundarylargelanguage}
Moxin Li, Yong Zhao, Yang Deng, Wenxuan Zhang, Shuaiyi Li, Wenya Xie, See-Kiong Ng, and Tat-Seng Chua.
\newblock Knowledge boundary of large language models: A survey, 2024{\natexlab{a}}.
\newblock URL \url{https://arxiv.org/abs/2412.12472}.

\bibitem[Li et~al.(2025{\natexlab{a}})Li, Xia, Yuan, Dong, Sha, Li, and Sui]{li2025digitaltwin}
Rui Li, Heming Xia, Xinfeng Yuan, Qingxiu Dong, Lei Sha, Wenjie Li, and Zhifang Sui.
\newblock How far are llms from being our digital twins? a benchmark for persona-based behavior chain simulation.
\newblock \emph{arXiv preprint arXiv:2502.14642}, 2025{\natexlab{a}}.

\bibitem[Li et~al.(2025{\natexlab{b}})Li, Li, Dong, Zhang, Zhang, Liu, Wang, Tang, and Liu]{li2025adaptive}
Wenjun Li, Dexun Li, Kuicai Dong, Cong Zhang, Hao Zhang, Weiwen Liu, Yasheng Wang, Ruiming Tang, and Yong Liu.
\newblock Adaptive tool use in large language models with meta-cognition trigger.
\newblock \emph{arXiv preprint arXiv:2502.12961}, 2025{\natexlab{b}}.

\bibitem[Li et~al.(2024{\natexlab{b}})Li, Chen, Ross, Huber, Moon, Lin, Dong, Sagar, Yan, and Crook]{li2024fnctod}
Zekun Li, Zhiyu Chen, Mike Ross, Patrick Huber, Seungwhan Moon, Zhaojiang Lin, Xin Dong, Adithya Sagar, Xifeng Yan, and Paul Crook.
\newblock Large language models as zero-shot dialogue state tracker through function calling.
\newblock In Lun-Wei Ku, Andre Martins, and Vivek Srikumar (eds.), \emph{Proceedings of the 62nd Annual Meeting of the Association for Computational Linguistics (Volume 1: Long Papers)}, pp.\  8688--8704, Bangkok, Thailand, August 2024{\natexlab{b}}. Association for Computational Linguistics.
\newblock \doi{10.18653/v1/2024.acl-long.471}.
\newblock URL \url{https://aclanthology.org/2024.acl-long.471/}.

\bibitem[Li et~al.(2025{\natexlab{c}})Li, Zhang, Zhang, Zhang, Liu, Yao, Xu, Zheng, Wang, Chen, et~al.]{li2025system1}
Zhong-Zhi Li, Duzhen Zhang, Ming-Liang Zhang, Jiaxin Zhang, Zengyan Liu, Yuxuan Yao, Haotian Xu, Junhao Zheng, Pei-Jie Wang, Xiuyi Chen, et~al.
\newblock From system 1 to system 2: A survey of reasoning large language models.
\newblock \emph{arXiv preprint arXiv:2502.17419}, 2025{\natexlab{c}}.

\bibitem[Lightman et~al.(2024)Lightman, Kosaraju, Burda, Edwards, Baker, Lee, Leike, Schulman, Sutskever, and Cobbe]{lightman2024lvsbs}
Hunter Lightman, Vineet Kosaraju, Yuri Burda, Harrison Edwards, Bowen Baker, Teddy Lee, Jan Leike, John Schulman, Ilya Sutskever, and Karl Cobbe.
\newblock Let's verify step by step.
\newblock In \emph{The Twelfth International Conference on Learning Representations}, 2024.
\newblock URL \url{https://openreview.net/forum?id=v8L0pN6EOi}.

\bibitem[Lim et~al.(2023)Lim, Kang, Kim, Kim, Hur, and Lim]{lim2023beyond}
Jungwoo Lim, Myunghoon Kang, Jinsung Kim, Jeongwook Kim, Yuna Hur, and Heui-Seok Lim.
\newblock Beyond candidates: adaptive dialogue agent utilizing persona and knowledge.
\newblock In \emph{Findings of the Association for Computational Linguistics: EMNLP 2023}, pp.\  7950--7963, 2023.

\bibitem[Lin et~al.(2025)Lin, Wen, Peng, Nie, Liao, Wang, Mo, Zhou, Cheng, Zhao, Wang, and Zhang]{lin2025hammer}
Qiqiang Lin, Muning Wen, Qiuying Peng, Guanyu Nie, Junwei Liao, Jun Wang, Xiaoyun Mo, Jiamu Zhou, Cheng Cheng, Yin Zhao, Jun Wang, and Weinan Zhang.
\newblock Robust function-calling for on-device language model via function masking.
\newblock In \emph{The Thirteenth International Conference on Learning Representations}, 2025.
\newblock URL \url{https://openreview.net/forum?id=yVQcr4qjD6}.

\bibitem[Liu et~al.(2025{\natexlab{a}})Liu, Li, Zhang, Wang, He, Hong, Liu, Zhang, Song, Zhu, et~al.]{liu2025acognitive}
Bang Liu, Xinfeng Li, Jiayi Zhang, Jinlin Wang, Tanjin He, Sirui Hong, Hongzhang Liu, Shaokun Zhang, Kaitao Song, Kunlun Zhu, et~al.
\newblock Advances and challenges in foundation agents: From brain-inspired intelligence to evolutionary, collaborative, and safe systems.
\newblock \emph{arXiv preprint arXiv:2504.01990}, 2025{\natexlab{a}}.

\bibitem[Liu \& Lane(2017)Liu and Lane]{bing2017policy}
Bing Liu and Ian Lane.
\newblock Iterative policy learning in end-to-end trainable task-oriented neural dialog models.
\newblock In \emph{2017 IEEE Automatic Speech Recognition and Understanding Workshop (ASRU)}, pp.\  482--489, 2017.
\newblock \doi{10.1109/ASRU.2017.8268975}.

\bibitem[Liu et~al.(2016)Liu, Lowe, Serban, Noseworthy, Charlin, and Pineau]{liu2016not}
Chia-Wei Liu, Ryan Lowe, Iulian~V Serban, Michael Noseworthy, Laurent Charlin, and Joelle Pineau.
\newblock How not to evaluate your dialogue system: An empirical study of unsupervised evaluation metrics for dialogue response generation.
\newblock \emph{arXiv preprint arXiv:1603.08023}, 2016.

\bibitem[Liu et~al.(2021)Liu, Zheng, Demasi, Sabour, Li, Yu, Jiang, and Huang]{liu2021towards}
Siyang Liu, Chujie Zheng, Orianna Demasi, Sahand Sabour, Yu~Li, Zhou Yu, Yong Jiang, and Minlie Huang.
\newblock Towards emotional support dialog systems.
\newblock In \emph{Proceedings of the 59th Annual Meeting of the Association for Computational Linguistics and the 11th International Joint Conference on Natural Language Processing (Volume 1: Long Papers)}, pp.\  3469--3483, 2021.

\bibitem[Liu et~al.(2025{\natexlab{b}})Liu, Zeng, Huang, xinlong hao, Yu, Li, Wang, Gan, Liu, Yu, WANG, Wang, Ning, Hou, Wang, Wu, Xinzhi, Liu, Wang, Tang, Tu, Shang, Jiang, Tang, Lian, Liu, and Chen]{liu2025toolace}
Weiwen Liu, Xingshan Zeng, Xu~Huang, xinlong hao, Shuai Yu, Dexun Li, Shuai Wang, Weinan Gan, Zhengying Liu, Yuanqing Yu, Zezhong WANG, Yuxian Wang, Wu~Ning, Yutai Hou, Bin Wang, Chuhan Wu, Wang Xinzhi, Yong Liu, Yasheng Wang, Duyu Tang, Dandan Tu, Lifeng Shang, Xin Jiang, Ruiming Tang, Defu Lian, Qun Liu, and Enhong Chen.
\newblock Tool{ACE}: Enhancing function calling with accuracy, complexity, and diversity.
\newblock In \emph{The Thirteenth International Conference on Learning Representations}, 2025{\natexlab{b}}.
\newblock URL \url{https://openreview.net/forum?id=8EB8k6DdCU}.

\bibitem[Lu et~al.(2024)Lu, Yang, Qian, Chen, Luo, Wu, Wang, Cong, Zhang, Lin, et~al.]{lu2024proactive}
Yaxi Lu, Shenzhi Yang, Cheng Qian, Guirong Chen, Qinyu Luo, Yesai Wu, Huadong Wang, Xin Cong, Zhong Zhang, Yankai Lin, et~al.
\newblock Proactive agent: Shifting llm agents from reactive responses to active assistance.
\newblock \emph{arXiv preprint arXiv:2410.12361}, 2024.

\bibitem[Ma et~al.(2024)Ma, Song, Zhuang, Hao, and King]{ma2024surveyembodied}
Yueen Ma, Zixing Song, Yuzheng Zhuang, Jianye Hao, and Irwin King.
\newblock A survey on vision-language-action models for embodied ai.
\newblock \emph{arXiv preprint arXiv:2405.14093}, 2024.

\bibitem[Madaan et~al.(2023)Madaan, Tandon, Gupta, Hallinan, Gao, Wiegreffe, Alon, Dziri, Prabhumoye, Yang, et~al.]{madaan2023sr}
Aman Madaan, Niket Tandon, Prakhar Gupta, Skyler Hallinan, Luyu Gao, Sarah Wiegreffe, Uri Alon, Nouha Dziri, Shrimai Prabhumoye, Yiming Yang, et~al.
\newblock Self-refine: Iterative refinement with self-feedback.
\newblock \emph{Advances in Neural Information Processing Systems}, 36:\penalty0 46534--46594, 2023.

\bibitem[Madaan et~al.(2024)Madaan, Tandon, Gupta, Hallinan, Gao, Wiegreffe, Alon, Dziri, Prabhumoye, Yang, et~al.]{madaan2024selfrefine}
Aman Madaan, Niket Tandon, Prakhar Gupta, Skyler Hallinan, Luyu Gao, Sarah Wiegreffe, Uri Alon, Nouha Dziri, Shrimai Prabhumoye, Yiming Yang, et~al.
\newblock Self-refine: Iterative refinement with self-feedback.
\newblock \emph{Advances in Neural Information Processing Systems}, 36, 2024.

\bibitem[Mielke et~al.(2022)Mielke, Szlam, Dinan, and Boureau]{mielke2022reducingconvagentcalibration}
Sabrina~J Mielke, Arthur Szlam, Emily Dinan, and Y-Lan Boureau.
\newblock Reducing conversational agents’ overconfidence through linguistic calibration.
\newblock \emph{Transactions of the Association for Computational Linguistics}, 10:\penalty0 857--872, 2022.

\bibitem[Minsky(1986)]{Minsky1986}
Marvin Minsky.
\newblock \emph{The Society of Mind}.
\newblock Simon \& Schuster, 1986.

\bibitem[Mondorf \& Plank(2024)Mondorf and Plank]{mondorf2024beyond}
Philipp Mondorf and Barbara Plank.
\newblock Beyond accuracy: Evaluating the reasoning behavior of large language models - a survey.
\newblock In \emph{First Conference on Language Modeling}, 2024.
\newblock URL \url{https://openreview.net/forum?id=Lmjgl2n11u}.

\bibitem[Muennighoff et~al.(2025)Muennighoff, Yang, Shi, Li, Fei-Fei, Hajishirzi, Zettlemoyer, Liang, Cand{\`e}s, and Hashimoto]{muennighoff2025s1}
Niklas Muennighoff, Zitong Yang, Weijia Shi, Xiang~Lisa Li, Li~Fei-Fei, Hannaneh Hajishirzi, Luke Zettlemoyer, Percy Liang, Emmanuel Cand{\`e}s, and Tatsunori Hashimoto.
\newblock s1: Simple test-time scaling.
\newblock \emph{arXiv preprint arXiv:2501.19393}, 2025.

\bibitem[Ni et~al.(2022)Ni, Young, Pandelea, Xue, and Cambria]{ni2022recentadvancesdeeplearning}
Jinjie Ni, Tom Young, Vlad Pandelea, Fuzhao Xue, and Erik Cambria.
\newblock Recent advances in deep learning based dialogue systems: A systematic survey, 2022.
\newblock URL \url{https://arxiv.org/abs/2105.04387}.

\bibitem[Niu et~al.(2024)Niu, Wang, Cheng, Song, and Zhang]{niu2024enhancingdialoguestatetracking}
Cheng Niu, Xingguang Wang, Xuxin Cheng, Juntong Song, and Tong Zhang.
\newblock Enhancing dialogue state tracking models through llm-backed user-agents simulation, 2024.
\newblock URL \url{https://arxiv.org/abs/2405.13037}.

\bibitem[Paranjape et~al.(2023)Paranjape, Lundberg, Singh, Hajishirzi, Zettlemoyer, and Ribeiro]{paranjape2023art}
Bhargavi Paranjape, Scott Lundberg, Sameer Singh, Hannaneh Hajishirzi, Luke Zettlemoyer, and Marco~Tulio Ribeiro.
\newblock Art: Automatic multi-step reasoning and tool-use for large language models.
\newblock \emph{arXiv preprint arXiv:2303.09014}, 2023.

\bibitem[Parisi et~al.(2022)Parisi, Zhao, and Fiedel]{parisi2022talm}
Aaron Parisi, Yao Zhao, and Noah Fiedel.
\newblock Talm: Tool augmented language models.
\newblock \emph{arXiv preprint arXiv:2205.12255}, 2022.

\bibitem[Park et~al.(2023)Park, O'Brien, Cai, Morris, Liang, and Bernstein]{generative_agents}
Joon~Sung Park, Joseph O'Brien, Carrie~Jun Cai, Meredith~Ringel Morris, Percy Liang, and Michael~S. Bernstein.
\newblock Generative agents: Interactive simulacra of human behavior.
\newblock In \emph{Proceedings of the 36th Annual ACM Symposium on User Interface Software and Technology}, UIST '23, New York, NY, USA, 2023. Association for Computing Machinery.
\newblock ISBN 9798400701320.
\newblock \doi{10.1145/3586183.3606763}.
\newblock URL \url{https://doi.org/10.1145/3586183.3606763}.

\bibitem[Patil et~al.(2023)Patil, Zhang, Wang, and Gonzalez]{patil2023gorilla}
Shishir~G Patil, Tianjun Zhang, Xin Wang, and Joseph~E Gonzalez.
\newblock Gorilla: Large language model connected with massive apis.
\newblock \emph{arXiv preprint arXiv:2305.15334}, 2023.

\bibitem[Peng et~al.(2018)Peng, Li, Gao, Liu, and Wong]{peng-etal-2018-deep}
Baolin Peng, Xiujun Li, Jianfeng Gao, Jingjing Liu, and Kam-Fai Wong.
\newblock {D}eep {D}yna-{Q}: Integrating planning for task-completion dialogue policy learning.
\newblock In Iryna Gurevych and Yusuke Miyao (eds.), \emph{Proceedings of the 56th Annual Meeting of the Association for Computational Linguistics (Volume 1: Long Papers)}, pp.\  2182--2192, Melbourne, Australia, July 2018. Association for Computational Linguistics.
\newblock \doi{10.18653/v1/P18-1203}.
\newblock URL \url{https://aclanthology.org/P18-1203/}.

\bibitem[Peng et~al.(2025)Peng, Qi, Wang, Yao, Xu, Hou, and Li]{peng2025agentic}
Hao Peng, Yunjia Qi, Xiaozhi Wang, Zijun Yao, Bin Xu, Lei Hou, and Juanzi Li.
\newblock Agentic reward modeling: Integrating human preferences with verifiable correctness signals for reliable reward systems.
\newblock \emph{arXiv preprint arXiv:2502.19328}, 2025.

\bibitem[Qian et~al.(2023)Qian, Han, Fung, Qin, Liu, and Ji]{qian2023creator}
Cheng Qian, Chi Han, Yi~Fung, Yujia Qin, Zhiyuan Liu, and Heng Ji.
\newblock Creator: Tool creation for disentangling abstract and concrete reasoning of large language models.
\newblock In \emph{Findings of the Association for Computational Linguistics: EMNLP 2023}, pp.\  6922--6939, 2023.

\bibitem[Qian et~al.(2024)Qian, He, Zhuang, Deng, Qin, Cong, Zhang, Zhou, Lin, Liu, and Sun]{qian2024tell}
Cheng Qian, Bingxiang He, Zhong Zhuang, Jia Deng, Yujia Qin, Xin Cong, Zhong Zhang, Jie Zhou, Yankai Lin, Zhiyuan Liu, and Maosong Sun.
\newblock Tell me more! towards implicit user intention understanding of language model driven agents.
\newblock In Lun-Wei Ku, Andre Martins, and Vivek Srikumar (eds.), \emph{Proceedings of the 62nd Annual Meeting of the Association for Computational Linguistics (Volume 1: Long Papers)}, pp.\  1088--1113, Bangkok, Thailand, August 2024. Association for Computational Linguistics.
\newblock \doi{10.18653/v1/2024.acl-long.61}.
\newblock URL \url{https://aclanthology.org/2024.acl-long.61}.

\bibitem[Qian et~al.(2025)Qian, Acikgoz, Wang, Chen, Sil, Hakkani-T{\"u}r, Tur, and Ji]{qian2025smart}
Cheng Qian, Emre~Can Acikgoz, Hongru Wang, Xiusi Chen, Avirup Sil, Dilek Hakkani-T{\"u}r, Gokhan Tur, and Heng Ji.
\newblock Smart: Self-aware agent for tool overuse mitigation.
\newblock \emph{arXiv preprint arXiv:2502.11435}, 2025.

\bibitem[Qiao et~al.(2025)Qiao, Qiu, Ren, Wang, Ru, Zhang, Chen, Jiang, Xie, Huang, and Chen]{qiao2025agentic}
Shuofei Qiao, Zhisong Qiu, Baochang Ren, Xiaobin Wang, Xiangyuan Ru, Ningyu Zhang, Xiang Chen, Yong Jiang, Pengjun Xie, Fei Huang, and Huajun Chen.
\newblock Agentic knowledgeable self-awareness.
\newblock In \emph{Workshop on Reasoning and Planning for Large Language Models}, 2025.
\newblock URL \url{https://openreview.net/forum?id=PGdSLjYwMT}.

\bibitem[Qin et~al.(2024{\natexlab{a}})Qin, Hu, Lin, Chen, Ding, Cui, Zeng, Zhou, Huang, Xiao, Han, Fung, Su, Wang, Qian, Tian, Zhu, Liang, Shen, Xu, Zhang, Ye, Li, Tang, Yi, Zhu, Dai, Yan, Cong, Lu, Zhao, Huang, Yan, Han, Sun, Li, Phang, Yang, Wu, Ji, Li, Liu, and Sun]{qin2024toollearningfm}
Yujia Qin, Shengding Hu, Yankai Lin, Weize Chen, Ning Ding, Ganqu Cui, Zheni Zeng, Xuanhe Zhou, Yufei Huang, Chaojun Xiao, Chi Han, Yi~Ren Fung, Yusheng Su, Huadong Wang, Cheng Qian, Runchu Tian, Kunlun Zhu, Shihao Liang, Xingyu Shen, Bokai Xu, Zhen Zhang, Yining Ye, Bowen Li, Ziwei Tang, Jing Yi, Yuzhang Zhu, Zhenning Dai, Lan Yan, Xin Cong, Yaxi Lu, Weilin Zhao, Yuxiang Huang, Junxi Yan, Xu~Han, Xian Sun, Dahai Li, Jason Phang, Cheng Yang, Tongshuang Wu, Heng Ji, Guoliang Li, Zhiyuan Liu, and Maosong Sun.
\newblock Tool learning with foundation models.
\newblock \emph{ACM Comput. Surv.}, 57\penalty0 (4), December 2024{\natexlab{a}}.
\newblock ISSN 0360-0300.
\newblock \doi{10.1145/3704435}.
\newblock URL \url{https://doi.org/10.1145/3704435}.

\bibitem[Qin et~al.(2024{\natexlab{b}})Qin, Liang, Ye, Zhu, Yan, Lu, Lin, Cong, Tang, Qian, Zhao, Hong, Tian, Xie, Zhou, Gerstein, dahai li, Liu, and Sun]{qin2024toolllm}
Yujia Qin, Shihao Liang, Yining Ye, Kunlun Zhu, Lan Yan, Yaxi Lu, Yankai Lin, Xin Cong, Xiangru Tang, Bill Qian, Sihan Zhao, Lauren Hong, Runchu Tian, Ruobing Xie, Jie Zhou, Mark Gerstein, dahai li, Zhiyuan Liu, and Maosong Sun.
\newblock Tool{LLM}: Facilitating large language models to master 16000+ real-world {API}s.
\newblock In \emph{The Twelfth International Conference on Learning Representations}, 2024{\natexlab{b}}.
\newblock URL \url{https://openreview.net/forum?id=dHng2O0Jjr}.

\bibitem[Qu et~al.(2025)Qu, Dai, Wei, Cai, Wang, Yin, Xu, and Wen]{qu2025tool}
Changle Qu, Sunhao Dai, Xiaochi Wei, Hengyi Cai, Shuaiqiang Wang, Dawei Yin, Jun Xu, and Ji-Rong Wen.
\newblock Tool learning with large language models: A survey.
\newblock \emph{Frontiers of Computer Science}, 19\penalty0 (8):\penalty0 198343, 2025.

\bibitem[Rashkin et~al.(2019)Rashkin, Smith, Li, and Boureau]{rashkin-etal-2019-towards}
Hannah Rashkin, Eric~Michael Smith, Margaret Li, and Y-Lan Boureau.
\newblock Towards empathetic open-domain conversation models: A new benchmark and dataset.
\newblock In Anna Korhonen, David Traum, and Llu{\'i}s M{\`a}rquez (eds.), \emph{Proceedings of the 57th Annual Meeting of the Association for Computational Linguistics}, pp.\  5370--5381, Florence, Italy, July 2019. Association for Computational Linguistics.
\newblock \doi{10.18653/v1/P19-1534}.
\newblock URL \url{https://aclanthology.org/P19-1534/}.

\bibitem[Sainz et~al.(2023)Sainz, Campos, Garc{\'i}a-Ferrero, Etxaniz, de~Lacalle, and Agirre]{sainz2023nlpcontamination}
Oscar Sainz, Jon Campos, Iker Garc{\'i}a-Ferrero, Julen Etxaniz, Oier~Lopez de~Lacalle, and Eneko Agirre.
\newblock {NLP} evaluation in trouble: On the need to measure {LLM} data contamination for each benchmark.
\newblock In Houda Bouamor, Juan Pino, and Kalika Bali (eds.), \emph{Findings of the Association for Computational Linguistics: EMNLP 2023}, pp.\  10776--10787, Singapore, December 2023. Association for Computational Linguistics.
\newblock \doi{10.18653/v1/2023.findings-emnlp.722}.
\newblock URL \url{https://aclanthology.org/2023.findings-emnlp.722/}.

\bibitem[Salemi et~al.(2024)Salemi, Mysore, Bendersky, and Zamani]{salemi2024lamp}
Alireza Salemi, Sheshera Mysore, Michael Bendersky, and Hamed Zamani.
\newblock Lamp: When large language models meet personalization.
\newblock In \emph{Proceedings of the 62nd Annual Meeting of the Association for Computational Linguistics (Volume 1: Long Papers)}, pp.\  7370--7392, 2024.

\bibitem[Schick et~al.(2023)Schick, Dwivedi-Yu, Dess{\`\i}, Raileanu, Lomeli, Hambro, Zettlemoyer, Cancedda, and Scialom]{schick2023toolformer}
Timo Schick, Jane Dwivedi-Yu, Roberto Dess{\`\i}, Roberta Raileanu, Maria Lomeli, Eric Hambro, Luke Zettlemoyer, Nicola Cancedda, and Thomas Scialom.
\newblock Toolformer: Language models can teach themselves to use tools.
\newblock \emph{Advances in Neural Information Processing Systems}, 36:\penalty0 68539--68551, 2023.

\bibitem[Shen et~al.(2023)Shen, Song, Tan, Li, Lu, and Zhuang]{shen2023hugginggpt}
Yongliang Shen, Kaitao Song, Xu~Tan, Dongsheng Li, Weiming Lu, and Yueting Zhuang.
\newblock Hugginggpt: Solving ai tasks with chatgpt and its friends in hugging face.
\newblock \emph{Advances in Neural Information Processing Systems}, 36:\penalty0 38154--38180, 2023.

\bibitem[Shinn et~al.(2023)Shinn, Cassano, Gopinath, Narasimhan, and Yao]{shinn2023reflexion}
Noah Shinn, Federico Cassano, Ashwin Gopinath, Karthik Narasimhan, and Shunyu Yao.
\newblock Reflexion: Language agents with verbal reinforcement learning.
\newblock \emph{Advances in Neural Information Processing Systems}, 36:\penalty0 8634--8652, 2023.

\bibitem[Shinn et~al.(2024)Shinn, Cassano, Gopinath, Narasimhan, and Yao]{shinn2024reflexion}
Noah Shinn, Federico Cassano, Ashwin Gopinath, Karthik Narasimhan, and Shunyu Yao.
\newblock Reflexion: Language agents with verbal reinforcement learning.
\newblock \emph{Advances in Neural Information Processing Systems}, 36, 2024.

\bibitem[Shridhar et~al.(2021)Shridhar, Yuan, Cote, Bisk, Trischler, and Hausknecht]{shridhar2021alfworld}
Mohit Shridhar, Xingdi Yuan, Marc-Alexandre Cote, Yonatan Bisk, Adam Trischler, and Matthew Hausknecht.
\newblock {\{}ALFW{\}}orld: Aligning text and embodied environments for interactive learning.
\newblock In \emph{International Conference on Learning Representations}, 2021.
\newblock URL \url{https://openreview.net/forum?id=0IOX0YcCdTn}.

\bibitem[Song et~al.(2024)Song, Yin, Yue, Huang, Li, and Lin]{song2024eto}
Yifan Song, Da~Yin, Xiang Yue, Jie Huang, Sujian Li, and Bill~Yuchen Lin.
\newblock Trial and error: Exploration-based trajectory optimization of llm agents.
\newblock In \emph{Proceedings of the 62nd Annual Meeting of the Association for Computational Linguistics (Volume 1: Long Papers)}, pp.\  7584--7600, 2024.

\bibitem[Srinivasan et~al.(2023)Srinivasan, Dong, Zhu, Yu, Mao, Mosk-Aoyama, Keutzer, Jiao, and Zhang]{srinivasan2023nexusraven}
Venkat~Krishna Srinivasan, Zhen Dong, Banghua Zhu, Brian Yu, Hanzi Mao, Damon Mosk-Aoyama, Kurt Keutzer, Jiantao Jiao, and Jian Zhang.
\newblock Nexusraven: a commercially-permissive language model for function calling.
\newblock In \emph{NeurIPS 2023 Workshop on Instruction Tuning and Instruction Following}, 2023.
\newblock URL \url{https://openreview.net/forum?id=Md6RUrGz67}.

\bibitem[Su et~al.(2023)Su, Xie, Huang, Song, Fang, Huang, and Feng]{su-etal-2023-scalable}
Haoxiang Su, Hongyan Xie, Hao Huang, Shuangyong Song, Ruiyu Fang, Xiaomeng Huang, and Sijie Feng.
\newblock Scalable-{DSC}: A structural template prompt approach to scalable dialogue state correction.
\newblock In Houda Bouamor, Juan Pino, and Kalika Bali (eds.), \emph{Proceedings of the 2023 Conference on Empirical Methods in Natural Language Processing}, pp.\  7902--7914, Singapore, December 2023. Association for Computational Linguistics.
\newblock \doi{10.18653/v1/2023.emnlp-main.490}.
\newblock URL \url{https://aclanthology.org/2023.emnlp-main.490/}.

\bibitem[Sumers et~al.(2024)Sumers, Yao, Narasimhan, and Griffiths]{sumers2024cognitive}
Theodore Sumers, Shunyu Yao, Karthik Narasimhan, and Thomas Griffiths.
\newblock Cognitive architectures for language agents.
\newblock \emph{Transactions on Machine Learning Research}, 2024.
\newblock ISSN 2835-8856.
\newblock URL \url{https://openreview.net/forum?id=1i6ZCvflQJ}.
\newblock Survey Certification.

\bibitem[Tan et~al.(2024{\natexlab{a}})Tan, Liu, and Jiang]{tan2024perpcs}
Zhaoxuan Tan, Zheyuan Liu, and Meng Jiang.
\newblock Personalized pieces: Efficient personalized large language models through collaborative efforts.
\newblock In Yaser Al-Onaizan, Mohit Bansal, and Yun-Nung Chen (eds.), \emph{Proceedings of the 2024 Conference on Empirical Methods in Natural Language Processing}, pp.\  6459--6475, Miami, Florida, USA, November 2024{\natexlab{a}}. Association for Computational Linguistics.
\newblock \doi{10.18653/v1/2024.emnlp-main.371}.
\newblock URL \url{https://aclanthology.org/2024.emnlp-main.371/}.

\bibitem[Tan et~al.(2024{\natexlab{b}})Tan, Zeng, Tian, Liu, Yin, and Jiang]{tan2024oppu}
Zhaoxuan Tan, Qingkai Zeng, Yijun Tian, Zheyuan Liu, Bing Yin, and Meng Jiang.
\newblock Democratizing large language models via personalized parameter-efficient fine-tuning.
\newblock In Yaser Al-Onaizan, Mohit Bansal, and Yun-Nung Chen (eds.), \emph{Proceedings of the 2024 Conference on Empirical Methods in Natural Language Processing}, pp.\  6476--6491, Miami, Florida, USA, November 2024{\natexlab{b}}. Association for Computational Linguistics.
\newblock \doi{10.18653/v1/2024.emnlp-main.372}.
\newblock URL \url{https://aclanthology.org/2024.emnlp-main.372/}.

\bibitem[Tang et~al.(2023)Tang, Deng, Lin, Han, Liang, Cao, and Sun]{tang2023toolalpaca}
Qiaoyu Tang, Ziliang Deng, Hongyu Lin, Xianpei Han, Qiao Liang, Boxi Cao, and Le~Sun.
\newblock Toolalpaca: Generalized tool learning for language models with 3000 simulated cases.
\newblock \emph{arXiv preprint arXiv:2306.05301}, 2023.

\bibitem[Team(2023)]{xagent2023}
XAgent Team.
\newblock Xagent: An autonomous agent for complex task solving, 2023.

\bibitem[Tu et~al.(2023)Tu, Chen, Li, Li, Shang, Zhao, Wang, and Yan]{tu2023characterchat}
Quan Tu, Chuanqi Chen, Jinpeng Li, Yanran Li, Shuo Shang, Dongyan Zhao, Ran Wang, and Rui Yan.
\newblock Characterchat: Learning towards conversational ai with personalized social support.
\newblock \emph{arXiv preprint arXiv:2308.10278}, 2023.

\bibitem[Ultes \& Maier(2021)Ultes and Maier]{ultes2021blending}
Stefan Ultes and Wolfgang Maier.
\newblock Blending task success and user satisfaction: Analysis of learned dialogue behaviour with multiple rewards.
\newblock In \emph{Proceedings of the 22nd Annual Meeting of the Special Interest Group on Discourse and Dialogue}, pp.\  403--410, 2021.

\bibitem[Wang et~al.(2021)Wang, Lin, Liu, and Wong]{wang-etal-2021-fast}
Dingmin Wang, Chenghua Lin, Qi~Liu, and Kam-Fai Wong.
\newblock Fast and scalable dialogue state tracking with explicit modular decomposition.
\newblock In Kristina Toutanova, Anna Rumshisky, Luke Zettlemoyer, Dilek Hakkani-Tur, Iz~Beltagy, Steven Bethard, Ryan Cotterell, Tanmoy Chakraborty, and Yichao Zhou (eds.), \emph{Proceedings of the 2021 Conference of the North American Chapter of the Association for Computational Linguistics: Human Language Technologies}, pp.\  289--295, Online, June 2021. Association for Computational Linguistics.
\newblock \doi{10.18653/v1/2021.naacl-main.27}.
\newblock URL \url{https://aclanthology.org/2021.naacl-main.27/}.

\bibitem[Wang et~al.(2023{\natexlab{a}})Wang, Hu, Deng, Wang, Mi, Wang, Wang, Kwan, King, and Wong]{wang-etal-2023-large}
Hongru Wang, Minda Hu, Yang Deng, Rui Wang, Fei Mi, Weichao Wang, Yasheng Wang, Wai-Chung Kwan, Irwin King, and Kam-Fai Wong.
\newblock Large language models as source planner for personalized knowledge-grounded dialogues.
\newblock In Houda Bouamor, Juan Pino, and Kalika Bali (eds.), \emph{Findings of the Association for Computational Linguistics: EMNLP 2023}, pp.\  9556--9569, Singapore, December 2023{\natexlab{a}}. Association for Computational Linguistics.
\newblock \doi{10.18653/v1/2023.findings-emnlp.641}.
\newblock URL \url{https://aclanthology.org/2023.findings-emnlp.641/}.

\bibitem[Wang et~al.(2023{\natexlab{b}})Wang, Wang, Du, Chen, Zhou, Wang, and Wong]{wang2023ds_survey}
Hongru Wang, Lingzhi Wang, Yiming Du, Liang Chen, Jingyan Zhou, Yufei Wang, and Kam-Fai Wong.
\newblock A survey of the evolution of language model-based dialogue systems, 2023{\natexlab{b}}.
\newblock URL \url{https://arxiv.org/abs/2311.16789}.

\bibitem[Wang et~al.(2023{\natexlab{c}})Wang, Wang, Mi, Deng, Wang, Liang, Xu, and Wong]{wang-etal-2023-cue}
Hongru Wang, Rui Wang, Fei Mi, Yang Deng, Zezhong Wang, Bin Liang, Ruifeng Xu, and Kam-Fai Wong.
\newblock Cue-{C}o{T}: Chain-of-thought prompting for responding to in-depth dialogue questions with {LLM}s.
\newblock In Houda Bouamor, Juan Pino, and Kalika Bali (eds.), \emph{Findings of the Association for Computational Linguistics: EMNLP 2023}, pp.\  12047--12064, Singapore, December 2023{\natexlab{c}}. Association for Computational Linguistics.
\newblock \doi{10.18653/v1/2023.findings-emnlp.806}.
\newblock URL \url{https://aclanthology.org/2023.findings-emnlp.806/}.

\bibitem[Wang et~al.(2024{\natexlab{a}})Wang, Wang, Xue, Xia, Cao, Liu, Pan, and Wong]{wang-etal-2024-appbench}
Hongru Wang, Rui Wang, Boyang Xue, Heming Xia, Jingtao Cao, Zeming Liu, Jeff~Z. Pan, and Kam-Fai Wong.
\newblock {A}pp{B}ench: Planning of multiple {API}s from various {APP}s for complex user instruction.
\newblock In Yaser Al-Onaizan, Mohit Bansal, and Yun-Nung Chen (eds.), \emph{Proceedings of the 2024 Conference on Empirical Methods in Natural Language Processing}, pp.\  15322--15336, Miami, Florida, USA, November 2024{\natexlab{a}}. Association for Computational Linguistics.
\newblock \doi{10.18653/v1/2024.emnlp-main.856}.
\newblock URL \url{https://aclanthology.org/2024.emnlp-main.856/}.

\bibitem[Wang et~al.(2025)Wang, Xue, Zhou, Zhang, Wang, Wang, Chen, and fai Wong]{wang2025selfdcreasonactself}
Hongru Wang, Boyang Xue, Baohang Zhou, Tianhua Zhang, Cunxiang Wang, Huimin Wang, Guanhua Chen, and Kam fai Wong.
\newblock Self-dc: When to reason and when to act? self divide-and-conquer for compositional unknown questions, 2025.
\newblock URL \url{https://arxiv.org/abs/2402.13514}.

\bibitem[Wang et~al.(2024{\natexlab{b}})Wang, Ma, Feng, Zhang, Yang, Zhang, Chen, Tang, Chen, Lin, et~al.]{wang2024survey}
Lei Wang, Chen Ma, Xueyang Feng, Zeyu Zhang, Hao Yang, Jingsen Zhang, Zhiyuan Chen, Jiakai Tang, Xu~Chen, Yankai Lin, et~al.
\newblock A survey on large language model based autonomous agents.
\newblock \emph{Frontiers of Computer Science}, 18\penalty0 (6):\penalty0 186345, 2024{\natexlab{b}}.

\bibitem[Wang et~al.(2022{\natexlab{a}})Wang, Jansen, C{\^o}t{\'e}, and Ammanabrolu]{wang2022scienceworld}
Ruoyao Wang, Peter Jansen, Marc-Alexandre C{\^o}t{\'e}, and Prithviraj Ammanabrolu.
\newblock {S}cience{W}orld: Is your agent smarter than a 5th grader?
\newblock In Yoav Goldberg, Zornitsa Kozareva, and Yue Zhang (eds.), \emph{Proceedings of the 2022 Conference on Empirical Methods in Natural Language Processing}, pp.\  11279--11298, Abu Dhabi, United Arab Emirates, December 2022{\natexlab{a}}. Association for Computational Linguistics.
\newblock \doi{10.18653/v1/2022.emnlp-main.775}.
\newblock URL \url{https://aclanthology.org/2022.emnlp-main.775/}.

\bibitem[Wang et~al.(2024{\natexlab{c}})Wang, Shi, Wang, Lee, Yuan, Huang, and Lyu]{wang2024learning}
Wenxuan Wang, Juluan Shi, Chaozheng Wang, Cheryl Lee, Youliang Yuan, Jen-tse Huang, and Michael~R Lyu.
\newblock Learning to ask: When llms meet unclear instruction.
\newblock \emph{arXiv preprint arXiv:2409.00557}, 2024{\natexlab{c}}.

\bibitem[Wang et~al.(2024{\natexlab{d}})Wang, Chen, Yuan, Zhang, Li, Peng, and Ji]{wang2024codeact}
Xingyao Wang, Yangyi Chen, Lifan Yuan, Yizhe Zhang, Yunzhu Li, Hao Peng, and Heng Ji.
\newblock Executable code actions elicit better llm agents.
\newblock In \emph{International Conference on Machine Learning}, pp.\  50208--50232. PMLR, 2024{\natexlab{d}}.

\bibitem[Wang et~al.(2022{\natexlab{b}})Wang, Wei, Schuurmans, Le, Chi, Narang, Chowdhery, and Zhou]{wang2022sc}
Xuezhi Wang, Jason Wei, Dale Schuurmans, Quoc Le, Ed~Chi, Sharan Narang, Aakanksha Chowdhery, and Denny Zhou.
\newblock Self-consistency improves chain of thought reasoning in language models.
\newblock \emph{arXiv preprint arXiv:2203.11171}, 2022{\natexlab{b}}.

\bibitem[Wei et~al.(2022)Wei, Wang, Schuurmans, Bosma, Xia, Chi, Le, Zhou, et~al.]{wei2022chain}
Jason Wei, Xuezhi Wang, Dale Schuurmans, Maarten Bosma, Fei Xia, Ed~Chi, Quoc~V Le, Denny Zhou, et~al.
\newblock Chain-of-thought prompting elicits reasoning in large language models.
\newblock \emph{Advances in neural information processing systems}, 35:\penalty0 24824--24837, 2022.

\bibitem[Wen et~al.(2017)Wen, Vandyke, Mrk{\v{s}}i{\'c}, Ga{\v{s}}i{\'c}, Rojas-Barahona, Su, Ultes, and Young]{wen-etal-2017-network}
Tsung-Hsien Wen, David Vandyke, Nikola Mrk{\v{s}}i{\'c}, Milica Ga{\v{s}}i{\'c}, Lina~M. Rojas-Barahona, Pei-Hao Su, Stefan Ultes, and Steve Young.
\newblock A network-based end-to-end trainable task-oriented dialogue system.
\newblock In Mirella Lapata, Phil Blunsom, and Alexander Koller (eds.), \emph{Proceedings of the 15th Conference of the {E}uropean Chapter of the Association for Computational Linguistics: Volume 1, Long Papers}, pp.\  438--449, Valencia, Spain, April 2017. Association for Computational Linguistics.
\newblock URL \url{https://aclanthology.org/E17-1042/}.

\bibitem[Wu et~al.(2024{\natexlab{a}})Wu, Zhu, Han, Tan, Zhang, and Chen]{wu2024seal}
Mengsong Wu, Tong Zhu, Han Han, Chuanyuan Tan, Xiang Zhang, and Wenliang Chen.
\newblock Seal-tools: Self-instruct tool learning dataset for agent tuning and detailed benchmark.
\newblock In \emph{CCF International Conference on Natural Language Processing and Chinese Computing}, pp.\  372--384. Springer, 2024{\natexlab{a}}.

\bibitem[Wu et~al.(2024{\natexlab{b}})Wu, Bansal, Zhang, Wu, Li, Zhu, Jiang, Zhang, Zhang, Liu, Awadallah, White, Burger, and Wang]{wu2024autogen}
Qingyun Wu, Gagan Bansal, Jieyu Zhang, Yiran Wu, Beibin Li, Erkang Zhu, Li~Jiang, Xiaoyun Zhang, Shaokun Zhang, Jiale Liu, Ahmed~Hassan Awadallah, Ryen~W White, Doug Burger, and Chi Wang.
\newblock Autogen: Enabling next-gen {LLM} applications via multi-agent conversations.
\newblock In \emph{First Conference on Language Modeling}, 2024{\natexlab{b}}.
\newblock URL \url{https://openreview.net/forum?id=BAakY1hNKS}.

\bibitem[Xi et~al.(2025)Xi, Chen, Guo, He, Ding, Hong, Zhang, Wang, Jin, Zhou, et~al.]{xi2025risesurveyagents}
Zhiheng Xi, Wenxiang Chen, Xin Guo, Wei He, Yiwen Ding, Boyang Hong, Ming Zhang, Junzhe Wang, Senjie Jin, Enyu Zhou, et~al.
\newblock The rise and potential of large language model based agents: A survey.
\newblock \emph{Science China Information Sciences}, 68\penalty0 (2):\penalty0 121101, 2025.

\bibitem[Xiao et~al.(2024)Xiao, Ma, Wang, Wu, Zhao, Wang, Huang, and Li]{xiao2024flowbench}
Ruixuan Xiao, Wentao Ma, Ke~Wang, Yuchuan Wu, Junbo Zhao, Haobo Wang, Fei Huang, and Yongbin Li.
\newblock Flowbench: Revisiting and benchmarking workflow-guided planning for llm-based agents.
\newblock In \emph{Findings of the Association for Computational Linguistics: EMNLP 2024}, pp.\  10883--10900, 2024.

\bibitem[Xie et~al.(2024{\natexlab{a}})Xie, Zhang, Chen, Zhu, Lou, Tian, Xiao, and Su]{xie2024travelplanner}
Jian Xie, Kai Zhang, Jiangjie Chen, Tinghui Zhu, Renze Lou, Yuandong Tian, Yanghua Xiao, and Yu~Su.
\newblock Travelplanner: A benchmark for real-world planning with language agents.
\newblock In \emph{Forty-first International Conference on Machine Learning}, 2024{\natexlab{a}}.
\newblock URL \url{https://openreview.net/forum?id=l5XQzNkAOe}.

\bibitem[Xie et~al.(2024{\natexlab{b}})Xie, Chen, Zhang, Wan, and Li]{xie2024largemultimodalagents}
Junlin Xie, Zhihong Chen, Ruifei Zhang, Xiang Wan, and Guanbin Li.
\newblock Large multimodal agents: A survey.
\newblock \emph{arXiv preprint arXiv:2402.15116}, 2024{\natexlab{b}}.

\bibitem[Xiong et~al.(2024)Xiong, Song, Zhao, Wu, Wang, Wang, Li, Peng, and Li]{xiong2024ipr}
Weimin Xiong, Yifan Song, Xiutian Zhao, Wenhao Wu, Xun Wang, Ke~Wang, Cheng Li, Wei Peng, and Sujian Li.
\newblock Watch every step! {LLM} agent learning via iterative step-level process refinement.
\newblock In Yaser Al-Onaizan, Mohit Bansal, and Yun-Nung Chen (eds.), \emph{Proceedings of the 2024 Conference on Empirical Methods in Natural Language Processing}, pp.\  1556--1572, Miami, Florida, USA, November 2024. Association for Computational Linguistics.
\newblock \doi{10.18653/v1/2024.emnlp-main.93}.
\newblock URL \url{https://aclanthology.org/2024.emnlp-main.93/}.

\bibitem[Xu et~al.(2024)Xu, Mao, Yang, Sun, and Huang]{xu2024autotod}
Heng-Da Xu, Xian-Ling Mao, Puhai Yang, Fanshu Sun, and Heyan Huang.
\newblock Rethinking task-oriented dialogue systems: From complex modularity to zero-shot autonomous agent.
\newblock In Lun-Wei Ku, Andre Martins, and Vivek Srikumar (eds.), \emph{Proceedings of the 62nd Annual Meeting of the Association for Computational Linguistics (Volume 1: Long Papers)}, pp.\  2748--2763, Bangkok, Thailand, August 2024. Association for Computational Linguistics.
\newblock \doi{10.18653/v1/2024.acl-long.152}.
\newblock URL \url{https://aclanthology.org/2024.acl-long.152/}.

\bibitem[Yan et~al.(2024{\natexlab{a}})Yan, Mao, Ji, Zhang, Patil, Stoica, and Gonzalez]{bfclv3}
Fanjia Yan, Huanzhi Mao, Charlie Cheng-Jie Ji, Tianjun Zhang, Shishir~G. Patil, Ion Stoica, and Joseph~E. Gonzalez.
\newblock Berkeley function calling leaderboard.
\newblock 2024{\natexlab{a}}.

\bibitem[Yan et~al.(2024{\natexlab{b}})Yan, Zhu, Zheng, Liu, Cao, Jiang, and Xu]{yan2024talk}
Haoqiu Yan, Yongxin Zhu, Kai Zheng, Bing Liu, Haoyu Cao, Deqiang Jiang, and Linli Xu.
\newblock Talk with human-like agents: Empathetic dialogue through perceptible acoustic reception and reaction.
\newblock \emph{arXiv preprint arXiv:2406.12707}, 2024{\natexlab{b}}.

\bibitem[Yao et~al.(2022{\natexlab{a}})Yao, Shi, Zou, Dai, Wu, Chen, Wang, and Yu]{yao-etal-2022-d4}
Binwei Yao, Chao Shi, Likai Zou, Lingfeng Dai, Mengyue Wu, Lu~Chen, Zhen Wang, and Kai Yu.
\newblock D4: a {C}hinese dialogue dataset for depression-diagnosis-oriented chat.
\newblock In Yoav Goldberg, Zornitsa Kozareva, and Yue Zhang (eds.), \emph{Proceedings of the 2022 Conference on Empirical Methods in Natural Language Processing}, pp.\  2438--2459, Abu Dhabi, United Arab Emirates, December 2022{\natexlab{a}}. Association for Computational Linguistics.
\newblock \doi{10.18653/v1/2022.emnlp-main.156}.
\newblock URL \url{https://aclanthology.org/2022.emnlp-main.156/}.

\bibitem[Yao et~al.(2022{\natexlab{b}})Yao, Chen, Yang, and Narasimhan]{yao2022webshop}
Shunyu Yao, Howard Chen, John Yang, and Karthik Narasimhan.
\newblock Webshop: Towards scalable real-world web interaction with grounded language agents.
\newblock \emph{Advances in Neural Information Processing Systems}, 35:\penalty0 20744--20757, 2022{\natexlab{b}}.

\bibitem[Yao et~al.(2023{\natexlab{a}})Yao, Yu, Zhao, Shafran, Griffiths, Cao, and Narasimhan]{yao2023tree}
Shunyu Yao, Dian Yu, Jeffrey Zhao, Izhak Shafran, Tom Griffiths, Yuan Cao, and Karthik Narasimhan.
\newblock Tree of thoughts: Deliberate problem solving with large language models.
\newblock \emph{Advances in neural information processing systems}, 36:\penalty0 11809--11822, 2023{\natexlab{a}}.

\bibitem[Yao et~al.(2023{\natexlab{b}})Yao, Zhao, Yu, Du, Shafran, Narasimhan, and Cao]{yao2023reactsynergizingreasoningacting-react}
Shunyu Yao, Jeffrey Zhao, Dian Yu, Nan Du, Izhak Shafran, Karthik Narasimhan, and Yuan Cao.
\newblock React: Synergizing reasoning and acting in language models, 2023{\natexlab{b}}.
\newblock URL \url{https://arxiv.org/abs/2210.03629}.

\bibitem[Yao et~al.(2024)Yao, Shinn, Razavi, and Narasimhan]{Yao2024tau}
Shunyu Yao, Noah Shinn, Pedram Razavi, and Karthik Narasimhan.
\newblock $\tau$-bench: A benchmark for tool-agent-user interaction in real-world domains.
\newblock \emph{ArXiv}, abs/2406.12045, 2024.
\newblock URL \url{https://api.semanticscholar.org/CorpusID:270562578}.

\bibitem[Yin et~al.(2023)Yin, Sun, Guo, Wu, Qiu, and Huang]{yin2023selfaware}
Zhangyue Yin, Qiushi Sun, Qipeng Guo, Jiawen Wu, Xipeng Qiu, and Xuan-Jing Huang.
\newblock Do large language models know what they don’t know?
\newblock In \emph{Findings of the Association for Computational Linguistics: ACL 2023}, pp.\  8653--8665, 2023.

\bibitem[Young(2002)]{young02_icslp}
Steve Young.
\newblock Talking to machines (statistically speaking).
\newblock In \emph{7th International Conference on Spoken Language Processing (ICSLP 2002)}, pp.\  9--16, 2002.
\newblock \doi{10.21437/ICSLP.2002-2}.

\bibitem[Yuan et~al.(2025)Yuan, Chen, Xi, Ye, Du, and Chen]{yuan2025agentr}
Siyu Yuan, Zehui Chen, Zhiheng Xi, Junjie Ye, Zhengyin Du, and Jiecao Chen.
\newblock Agent-r: Training language model agents to reflect via iterative self-training.
\newblock \emph{arXiv preprint arXiv:2501.11425}, 2025.

\bibitem[Yuksekgonul et~al.(2024)Yuksekgonul, Bianchi, Boen, Liu, Huang, Guestrin, and Zou]{yuksekgonul2024textgrad}
Mert Yuksekgonul, Federico Bianchi, Joseph Boen, Sheng Liu, Zhi Huang, Carlos Guestrin, and James Zou.
\newblock Textgrad: Automatic" differentiation" via text.
\newblock \emph{arXiv preprint arXiv:2406.07496}, 2024.

\bibitem[Zeng et~al.(2024)Zeng, Liu, Lu, Wang, Liu, Dong, and Tang]{zeng2024agenttuning}
Aohan Zeng, Mingdao Liu, Rui Lu, Bowen Wang, Xiao Liu, Yuxiao Dong, and Jie Tang.
\newblock {A}gent{T}uning: Enabling generalized agent abilities for {LLM}s.
\newblock In Lun-Wei Ku, Andre Martins, and Vivek Srikumar (eds.), \emph{Findings of the Association for Computational Linguistics: ACL 2024}, pp.\  3053--3077, Bangkok, Thailand, August 2024. Association for Computational Linguistics.
\newblock \doi{10.18653/v1/2024.findings-acl.181}.
\newblock URL \url{https://aclanthology.org/2024.findings-acl.181/}.

\bibitem[Zhang et~al.(2025{\natexlab{a}})Zhang, Dai, Wu, Yang, Wang, Tang, and Liu]{zhang2025multiturnsurvey}
Chen Zhang, Xinyi Dai, Yaxiong Wu, Qu~Yang, Yasheng Wang, Ruiming Tang, and Yong Liu.
\newblock A survey on multi-turn interaction capabilities of large language models.
\newblock \emph{arXiv preprint arXiv:2501.09959}, 2025{\natexlab{a}}.

\bibitem[Zhang et~al.(2024{\natexlab{a}})Zhang, Lan, Rithesh, Liu, Yao, Tan, Hoang, Yang, Feng, Liu, et~al.]{zhang2024agentohana}
Jianguo Zhang, Tian Lan, RN~Rithesh, Zhiwei Liu, Weiran Yao, Juntao Tan, Thai~Quoc Hoang, Liangwei Yang, Yihao Feng, Zuxin Liu, et~al.
\newblock The agent ohana: Designing unified data and training pipeline for effective agent learning.
\newblock In \emph{ICLR 2024 Workshop on Large Language Model (LLM) Agents}, 2024{\natexlab{a}}.

\bibitem[Zhang et~al.(2024{\natexlab{b}})Zhang, Lan, Zhu, Liu, Hoang, Kokane, Yao, Tan, Prabhakar, Chen, et~al.]{zhang2024xlam}
Jianguo Zhang, Tian Lan, Ming Zhu, Zuxin Liu, Thai Hoang, Shirley Kokane, Weiran Yao, Juntao Tan, Akshara Prabhakar, Haolin Chen, et~al.
\newblock xlam: A family of large action models to empower ai agent systems.
\newblock \emph{CoRR}, 2024{\natexlab{b}}.

\bibitem[Zhang \& Choi(2023)Zhang and Choi]{zhang2023clarify}
Michael~JQ Zhang and Eunsol Choi.
\newblock Clarify when necessary: Resolving ambiguity through interaction with lms.
\newblock \emph{arXiv preprint arXiv:2311.09469}, 2023.

\bibitem[Zhang et~al.(2025{\natexlab{b}})Zhang, Lyu, Sun, Wang, Zhang, Guo, Wang, King, Liu, and Ma]{zhang2025whatwherehow}
Qiyuan Zhang, Fuyuan Lyu, Zexu Sun, Lei Wang, Weixu Zhang, Zhihan Guo, Yufei Wang, Irwin King, Xue Liu, and Chen Ma.
\newblock What, how, where, and how well? a survey on test-time scaling in large language models.
\newblock \emph{arXiv preprint arXiv:2503.24235}, 2025{\natexlab{b}}.

\bibitem[Zhang et~al.(2018)Zhang, Dinan, Urbanek, Szlam, Kiela, and Weston]{zhang2018personalizingdialogue}
Saizheng Zhang, Emily Dinan, Jack Urbanek, Arthur Szlam, Douwe Kiela, and Jason Weston.
\newblock Personalizing dialogue agents: {I} have a dog, do you have pets too?
\newblock In Iryna Gurevych and Yusuke Miyao (eds.), \emph{Proceedings of the 56th Annual Meeting of the Association for Computational Linguistics (Volume 1: Long Papers)}, pp.\  2204--2213, Melbourne, Australia, July 2018. Association for Computational Linguistics.
\newblock \doi{10.18653/v1/P18-1205}.
\newblock URL \url{https://aclanthology.org/P18-1205/}.

\bibitem[Zhang et~al.(2023)Zhang, Peng, Li, Zhou, and Meng]{zhang2023sgptod}
Xiaoying Zhang, Baolin Peng, Kun Li, Jingyan Zhou, and Helen Meng.
\newblock {SGP}-{TOD}: Building task bots effortlessly via schema-guided {LLM} prompting.
\newblock In Houda Bouamor, Juan Pino, and Kalika Bali (eds.), \emph{Findings of the Association for Computational Linguistics: EMNLP 2023}, pp.\  13348--13369, Singapore, December 2023. Association for Computational Linguistics.
\newblock \doi{10.18653/v1/2023.findings-emnlp.891}.
\newblock URL \url{https://aclanthology.org/2023.findings-emnlp.891/}.

\bibitem[Zhang et~al.(2025{\natexlab{c}})Zhang, Shen, Zheng, Wu, Zhang, Yan, Peng, Wang, and Lu]{zhang2025asktoact}
Xuan Zhang, Yongliang Shen, Zhe Zheng, Linjuan Wu, Wenqi Zhang, Yuchen Yan, Qiuying Peng, Jun Wang, and Weiming Lu.
\newblock Asktoact: Enhancing llms tool use via self-correcting clarification.
\newblock \emph{arXiv preprint arXiv:2503.01940}, 2025{\natexlab{c}}.

\bibitem[Zhao et~al.(2023)Zhao, Zhao, Lu, Wang, Tong, and Qin]{zhao2023chatgptemotion}
Weixiang Zhao, Yanyan Zhao, Xin Lu, Shilong Wang, Yanpeng Tong, and Bing Qin.
\newblock Is chatgpt equipped with emotional dialogue capabilities?
\newblock \emph{arXiv preprint arXiv:2304.09582}, 2023.

\bibitem[Zheng et~al.(2021)Zheng, Liu, Chen, Leng, and Huang]{zheng-etal-2021-comae}
Chujie Zheng, Yong Liu, Wei Chen, Yongcai Leng, and Minlie Huang.
\newblock {C}o{MAE}: A multi-factor hierarchical framework for empathetic response generation.
\newblock In Chengqing Zong, Fei Xia, Wenjie Li, and Roberto Navigli (eds.), \emph{Findings of the Association for Computational Linguistics: ACL-IJCNLP 2021}, pp.\  813--824, Online, August 2021. Association for Computational Linguistics.
\newblock \doi{10.18653/v1/2021.findings-acl.72}.
\newblock URL \url{https://aclanthology.org/2021.findings-acl.72/}.

\bibitem[Zheng et~al.(2023)Zheng, Sabour, Wen, Zhang, and Huang]{zheng-etal-2023-augesc}
Chujie Zheng, Sahand Sabour, Jiaxin Wen, Zheng Zhang, and Minlie Huang.
\newblock {A}ug{ESC}: Dialogue augmentation with large language models for emotional support conversation.
\newblock In Anna Rogers, Jordan Boyd-Graber, and Naoaki Okazaki (eds.), \emph{Findings of the Association for Computational Linguistics: ACL 2023}, pp.\  1552--1568, Toronto, Canada, July 2023. Association for Computational Linguistics.
\newblock \doi{10.18653/v1/2023.findings-acl.99}.
\newblock URL \url{https://aclanthology.org/2023.findings-acl.99/}.

\bibitem[Zhou et~al.(2023{\natexlab{a}})Zhou, Yan, Shlapentokh-Rothman, Wang, and Wang]{zhou2023lats}
Andy Zhou, Kai Yan, Michal Shlapentokh-Rothman, Haohan Wang, and Yu-Xiong Wang.
\newblock Language agent tree search unifies reasoning acting and planning in language models.
\newblock \emph{arXiv preprint arXiv:2310.04406}, 2023{\natexlab{a}}.

\bibitem[Zhou et~al.(2023{\natexlab{b}})Zhou, Sch{\"a}rli, Hou, Wei, Scales, Wang, Schuurmans, Cui, Bousquet, Le, and Chi]{zhou2023leasttomost}
Denny Zhou, Nathanael Sch{\"a}rli, Le~Hou, Jason Wei, Nathan Scales, Xuezhi Wang, Dale Schuurmans, Claire Cui, Olivier Bousquet, Quoc~V Le, and Ed~H. Chi.
\newblock Least-to-most prompting enables complex reasoning in large language models.
\newblock In \emph{The Eleventh International Conference on Learning Representations}, 2023{\natexlab{b}}.
\newblock URL \url{https://openreview.net/forum?id=WZH7099tgfM}.

\bibitem[Zhu et~al.(2025)Zhu, Du, Hong, Yang, Guo, Wang, Wang, Qian, Tang, Ji, et~al.]{zhu2025multiagentbench}
Kunlun Zhu, Hongyi Du, Zhaochen Hong, Xiaocheng Yang, Shuyi Guo, Zhe Wang, Zhenhailong Wang, Cheng Qian, Xiangru Tang, Heng Ji, et~al.
\newblock Multiagentbench: Evaluating the collaboration and competition of llm agents.
\newblock \emph{arXiv preprint arXiv:2503.01935}, 2025.

\bibitem[Zhuang et~al.(2023)Zhuang, Chen, Yu, Mitra, Bursztyn, Rossi, Sarkhel, and Zhang]{zhuang2023toolchain}
Yuchen Zhuang, Xiang Chen, Tong Yu, Saayan Mitra, Victor Bursztyn, Ryan~A Rossi, Somdeb Sarkhel, and Chao Zhang.
\newblock Toolchain*: Efficient action space navigation in large language models with a* search.
\newblock \emph{arXiv preprint arXiv:2310.13227}, 2023.

\end{thebibliography}
\bibliographystyle{colm2025_conference}

\clearpage
\newpage
\appendix
\section*{Appendix}

\section{How Do Conversational Agents Differ from Basic Agents?}
\label{app:ca}
While recent agents can certainly perform intricate reasoning and tool-based operations, they typically do not engage in real-time, user-centric dialogues or adapt their decision-making based on evolving user feedback. In contrast, Conversational Agents must integrate complex reasoning with continuous context-aware conversation loops, dynamically refining their actions to align with the user’s changing goals and constraints. As demonstrated by LLM-based frameworks equipped with plugins or tool use, these agents can integrate diverse external knowledge (e.g., web APIs, databases) while continuously monitoring the conversation and ensuring alignment with the user’s context and goals, ultimately enabling personalized and contextually rich interactions.

\begin{table}[!h]
\centering
\resizebox{\linewidth}{!}{
\begin{tabular}{clccccc}
\toprule
\textbf{Dimension} & \textbf{Categories} &
\multicolumn{2}{c}{\textbf{Dialogue Systems}} & \multicolumn{1}{c}{\textbf{Language Agent}} & \multicolumn{1}{c}{\textbf{Conversational Agent}} \\
\cmidrule(lr){3-4}
& & \textbf{Open-ended} & \textbf{Task-oriented} & & \\ \midrule

\multirow{2}{*}{\textbf{Reasoning}} 
& \textit{General Reasoning}
& \cmark 
& \cmark
& \multicolumn{1}{c}{\cmark}
& \multicolumn{1}{c}{\cmark} \\

& \textit{Agentic Reasoning}
& \xmark 
& \gmark
& \multicolumn{1}{c}{\cmark}
& \multicolumn{1}{c}{\cmark} \\ \midrule

\multirow{5}{*}{\textbf{\makecell{Monitor}}} 
& \textit{Self-Impose Capability}
& \xmark
& \gmark
& \multicolumn{1}{c}{\gmark}
& \multicolumn{1}{c}{\cmark} \\

& \textit{Self-Correction}
& \xmark
& \xmark
& \multicolumn{1}{c}{\cmark}
& \multicolumn{1}{c}{\cmark} \\

& \textit{User State Tracking}
& \gmark 
& \cmark
& \multicolumn{1}{c}{\xmark}
& \multicolumn{1}{c}{\cmark} \\

& \textit{Personalization \& Persona}
& \gmark 
& \cmark
& \multicolumn{1}{c}{\cmark}
& \multicolumn{1}{c}{\cmark} \\

& \textit{Emotion \& Sentiment}
& \cmark 
& \gmark
& \multicolumn{1}{c}{\xmark}
& \multicolumn{1}{c}{\cmark} \\\midrule

\multirow{3}{*}{\textbf{\makecell{Control}}} 
& \textit{Tool Selection}      
& \gmark 
& \cmark 
& \multicolumn{1}{c}{\cmark}
& \multicolumn{1}{c}{\cmark} \\

& \textit{Tool Execution}      
& \xmark 
& \cmark 
& \multicolumn{1}{c}{\cmark}
& \multicolumn{1}{c}{\cmark} \\

& \textit{Policy Following}      
& \gmark 
& \cmark 
& \multicolumn{1}{c}{\xmark}
& \multicolumn{1}{c}{\cmark} \\ \bottomrule
\end{tabular}
}
\caption{Comparison of capabilities among Dialogue Systems, Language Agents, and Conversational Agents as addressed (\cmark), partially addressed (\gmark), and not addressed (\xmark).}
\label{tab:comparison}
\end{table}
\vspace{-4mm}

\section{Additional Details on Reasoning, Monitor, and Control}

\vspace{-2mm}
\label{app:rmc}
\noindent\textbf{Reasoning.} Conversational Agents leverage advanced reasoning techniques to break down complex tasks, interpret user objectives, and plan a sequence of steps for successful completion of tasks. Beyond simple response generation, these systems can integrate multi-step logic chains or iteratively refine their own decisions, allowing them to reach more accurate conclusions over time. Some frameworks adopt more agentic approaches, blending reasoning and acting to handle dynamic user needs or unforeseen events. Additionally, by incorporating user feedback at each stage, Conversational Agents can clarify ambiguous requirements, adapt to new incoming information, and collaboratively refine their reasoning to deliver increasingly robust and personalized solutions.

\noindent\textbf{Monitor.} A core capability of Conversational Agents lies in tracking both internal and user-centric states. Internally, they maintain self-awareness by monitoring their own performance, constraints, and opportunities for self-correction when errors or oversights arise. Externally, they focus on user and interaction monitoring by maintaining an evolving representation of user context—from preferences and past interactions to emotional cues—to deliver personalized and empathetic engagement. Although some designs may include additional environment awareness or external context under proactive behaviors, the key objective remains user awareness: proactively addressing user needs, asking clarifying questions, or adjusting strategies when objectives shift or are ambiguous.

\noindent\textbf{Control.} Finally, Conversational Agents can invoke external resources and tools on demand. Rather than relying solely on static internal knowledge, they can call APIs or databases to retrieve up-to-date information—such as flight prices or product availability—and perform actions like booking a reservation or placing an order. By weaving tool usage seamlessly into the conversation, these systems preserve a natural dialogue flow while executing complex tasks. Furthermore, adherence to policies or guidelines ensures that actions taken align with user constraints and ethical considerations.

\section{Evaluation of Conversational Agents}
\label{app:eval-ca}

Although evaluating agents is beyond the scope of our paper, we would like to share some discussion points as supplementary material for evaluating Conversational Agents, specifically on: (i) \textit{Tool Utilization} and (ii) \textit{Conversational Task Completion}. We also provide a comparison of their features in Table \ref{tab:bechmarks}.

\begin{table*}[!h]
\centering
\resizebox{1.0\linewidth}{!}{
\begin{tabular}{l c c c c c }
\toprule
\textbf{Benchmarks}                               & \textbf{\# of Samples}  & \textbf{Tool Execution}  & \textbf{Multi-Step}  & \textbf{Multi-Turn}  & \textbf{Real API}  \\ \midrule
ALFWorld~\citep{shridhar2021alfworld}            & 274                     & \cmark                  & \xmark               & \cmark              & \xmark            \\
ScienceWorld~\citep{wang2022scienceworld}        & 1,800                   & \cmark                  & \xmark              & \cmark              & \xmark            \\
Webshop~\citep{yao2022webshop}                   & 1,211                   & \cmark                  & \xmark               & \cmark              & \xmark            \\
API-Bank~\citep{li-etal-2023-api}                & 314                     & \xmark                  & \xmark               & \cmark              & \xmark            \\
ToolAlpaca~\citep{tang2023toolalpaca}            & 3,938                   & \xmark                  & \cmark               & \xmark              & \xmark            \\
NexusRaven~\citep{srinivasan2023nexusraven}      & 318                     & \xmark                  & \xmark              & \xmark              & \xmark            \\
TravelPlanner~\citep{xie2024travelplanner}       & 1,225                   &  \cmark                 & \cmark               & \xmark             & \xmark            \\ 
AppBench~\citep{wang-etal-2024-appbench}                     & 800                    & \xmark                  & \cmark              & \xmark             & \cmark            \\ 
Sea-tools~\citep{wu2024seal}                     & 294                     & \xmark                  & \xmark              & \xmark             & \xmark            \\ 
$\tau$-Bench~\citep{Yao2024tau}                  & 165                     & \cmark                  & \cmark              & \cmark             & \xmark            \\ 
BFCL-V3~\citep{bfclv3}                           & 4,751                   & \xmark                  & \cmark              & \cmark             & \cmark            \\ \bottomrule
\end{tabular}
}
\caption{Comparison of recent benchmarks for evaluating Conversational Agents.}
\vspace{-1mm}
\label{tab:bechmarks}
\end{table*}

\subsection{Tool Utilization Benchmarks}
API-Bank~\citep{li-etal-2023-api} pioneered comprehensive benchmarking for tool-augmented LLMs by introducing hundreds of annotated multi-turn dialogues, making it one of the first benchmarks to systematically evaluate a language agent’s ability to plan and select appropriate API calls in context. Similarly, ToolAlpaca~\citep{tang2023toolalpaca} introduced a novel self-generated dataset comprising nearly 4,000 diverse tool-use cases across over 400 APIs, leveraging multi-agent simulation to enable generalized tool use. In contrast, the evaluation sets of NexusRaven~\citep{srinivasan2023nexusraven} and Seal-Tools~\citep{wu2024seal} primarily focus on assessing the single-turn function-calling capabilities of LLMs. More recently, BFCL V3~\citep{bfclv3} expanded these benchmarks to specifically evaluate multi-turn, multi-step tool use, including real-time APIs, making it one of the most comprehensive and challenging benchmarks for assessing capabilities of language agents in function calling scenarios.

\subsection{Conversational Task Completion}

Beyond tool utilization, task completion benchmarks evaluate Conversational Agents' multi-turn capabilities and action-taking skills needed to achieve user-driven goals in interactive, multi-step environments grounded in real-world tasks. ALFWorld~\citep{shridhar2021alfworld} bridges textual planning and embodied execution by aligning abstract reasoning in a text-based simulator with different action sequences (e.g, open the cabinet) in a 3D household environment and ScienceWorld~\citep{wang2022scienceworld} presents an interactive text-based laboratory environment that evaluates scientific reasoning at a fifth-grade level by requiring agents to perform experiments and explain outcomes. On the other hand, WebShop~\citep{yao2022webshop} introduces a large-scale web interaction environment where an agent must fulfill realistic shopping requests from the user by navigating a simulated e-commerce site with over a million products, using search and submit actions.  Unlike these approaches, TravelPlanner~\citep{xie2024travelplanner} introduces a benchmark for evaluating the multi-step planning abilities of LLMs in the travel domain, requiring agents to generate complete itineraries using a suite of tools and satisfy user constraints. According to the results, most LLMs perform poorly on this benchmark. However, one limitation is that while the benchmark may require multiple subsequent or parallel function calls, it lacks multi-turn interaction with users. Most notably, $\tau$-bench~\citep{Yao2024tau} integrates both realistic tool utilization, policy following and long-horizon, multi-turn dialogue with simulated users. This dual emphasis makes $\tau$-bench particularly well-suited for evaluating Conversational Agents, as it captures the interplay between natural language interaction and sequential decision-making in complex task-oriented settings.


\end{document}